\documentclass[letterpaper, preprint, paper,11pt]{AAS}	

\usepackage{amsmath, amssymb, bm}
\usepackage{subfigure}
\usepackage[colorlinks=true, pdfstartview=FitV, linkcolor=black, citecolor= black, urlcolor= black]{hyperref}
\usepackage{overcite}
\usepackage{footnpag}			      	
\usepackage{algorithm}
\usepackage{algpseudocode}
\usepackage{float}
\usepackage{booktabs}
\usepackage{siunitx}
\usepackage{multirow}
 
\usepackage{subcaption}
\PaperNumber{XX-XXX}

\begin{document}

\title{Memory-Efficient Meta-Reinforcement Learning for Adaptive Safety-Critical Control in Adversarial Spacecraft Proximity Operations}

\author{Alejandro Posadas-Nava\thanks{Researcher, AeroAstro, MIT, 77 Massachusetts Avenue, Cambridge, MA 02139-4307.},
\,
Richard Linares\thanks{Associate Professor and Rockwell International Career Development Professor, AeroAstro, MIT, 77 Massachusetts Avenue, Cambridge, MA 02139-4307.}, 
 Minduli Wijayatunga\thanks{ Assistant Professor, Department of Aerospace Engineering, University of Illinois, Urbana-Champaign, 104 S Wright St, Urbana, IL 61801.}
}

\maketitle{}

\begin{abstract}
Autonomous spacecraft rendezvous and proximity operations (RPO) require controllers that guarantee safety under thrust constraints while minimizing fuel expenditure. Input-constrained control barrier functions (ICCBFs) provide a control method for nonlinear systems with actuation constraints that construct a forward-invariant safe set. Previous work has shown that learning class-$\mathcal{K}$ functions defining the ICCBF recursion via meta reinforcement learning (meta-RL) yields a robust, non-greedy approach to safety-critical control in RPO. This paper extends that framework further by investigating the performance of three recurrent network architectures (Long Short Term Memory (LSTM),  Gated Recurrent Unit (GRU), Selective State Space Model (Mamba)) and two training algorithms 
(Proximal Policy Optimization (PPO) and Soft Actor Critic (SAC)) to identify the best setup for tuning ICCBF class-$\mathcal{K}$ functions via meta-RL. In addition to cooperative test cases, performance is evaluated in the presence of adversarial behavior where the target spacecraft behaves in a way that worsens the safety of the chaser spacecraft. Results indicate that state space models such as Mamba when used with PPO achieve superior task completion, safety, and fuel-savings compared to other architectures, across all cooperative and uncooperative scenarios tested. 
\end{abstract}

\section{Introduction}

Autonomous rendezvous and proximity operations (RPO) increasingly require a spacecraft to maneuver in close proximity to objects whose properties are uncertain and, in some settings, actively uncooperative. This requires the spacecraft controls to meet safety guarantees under bounded thrust and significant model uncertainty, conserve propellant, and execute within the limited computational budget available on flight hardware.


Classical guidance and control (G\&C) architectures that separate trajectory planning from safety enforcement provide only weak guarantees under these coupled constraints. These approaches are rooted in optimal control and mathematical programming~\cite{betts1998survey}. Indirect methods derived from Pontryagin's minimum principle yield fuel-optimal solutions with high precision, though their convergence is notoriously sensitive to initialization and problem scaling~\cite{wijayatunga2023scaling}. Direct methods based on convex and successive convex programming trade some optimality for reliability and speed~\cite{lu2013conic, malyuta2022convex, bernardini2023trustregion}, and have matured into computationally tractable pipelines for trajectory design and guidance in debris removal and servicing missions~\cite{wijayatunga2023adrdesign}, including end-to-end close-range rendezvous frameworks validated on hardware testbeds~\cite{wijayatunga2026cortex}. Receding-horizon schemes such as model predictive control (MPC) further embed constraint handling within online re-optimization~\cite{weiss2015mpc}. However, the guarantees these methods provide are certified only with respect to the assumed model. Their optimality and constraint satisfaction  can degrade under unmodeled dynamics, navigation error, and parameter uncertainty, and robustness must be recovered indirectly through conservative margins, disturbance bounding, or repeated replanning. Moreover, these formulations offer no principled mechanism to infer hidden physical parameters or anticipate non-cooperative target behavior from the observation history.

In contrast, reinforcement learning (RL) offers a complementary set of strengths. By training closed-loop policies over distributions of dynamics, disturbances, and sensing conditions, RL produces controllers that adapt online to conditions unseen at design time while requiring only inexpensive forward inference onboard. It has consequently been applied across the spacecraft G\&C spectrum, from six-degree-of-freedom planetary landing~\cite{gaudet2020sixdof} and robust interplanetary trajectory design~\cite{zavoli2021robust} to autonomous guidance for proximity operations~\cite{federici2021deep}. In the RPO context specifically, RL-based guidance has been shown to maintain target observability, safety margins, and low fuel consumption under angles-only navigation in far-range rendezvous~\cite{wijayatunga2025farrange}.  Meta-RL extends this further: by equipping the policy with memory, the agent implicitly performs system identification within an episode, adapting to hidden parameters and time-varying environments without explicit estimators~\cite{federici2022meta, fereoli2024meta}. However, RL by itself provides no formal guarantees. Constraints are typically encoded as reward penalties, so satisfaction is achieved only in expectation and only on the training distribution; a single constraint violation during a close-approach maneuver can be mission-ending. This shortcoming has spurred safe-RL mechanisms such as shielding~\cite{alshiekh2018shielding} and run-time assurance for spacecraft docking and inspection~\cite{dunlap2023rta, vanwijk2024inspection}, and more broadly motivates hybrid architectures in which a learned policy is paired with a certificate-bearing safety mechanism that retains hard guarantees regardless of what the policy outputs~\cite{wijayatunga2026twostage}.

Control Barrier Functions (CBFs) offer such a mechanism, enforcing forward invariance of a prescribed safe set through a real-time quadratic program (QP) that minimally alters a nominal command~\cite{ames2019controlbarrierfunctionstheory}. Input-Constrained CBFs (ICCBFs) extend this guarantee to systems with bounded actuation by recursively composing the barrier with a hierarchy of class-$\mathcal{K}$ functions to construct an input-admissible inner safe set~\cite{Agrawal_2021}. While well suited to address thrust limits and position-only constraints in RPO, conventional ICCBFs use a fixed class-$\mathcal{K}$ function hierarchy. This renders the filter conservative and myopic, as it can shrink the feasible set unnecessarily, expend excess fuel, and behave sub-optimally near constraint boundaries.

Recent work has addressed this limitation by parameterizing and \emph{learning} the class-$\mathcal{K}$ hierarchy. Alongside data-driven approaches that represent barrier and certificate functions with neural networks~\cite{dawson2023certificates}, a first line of work demonstrated that RL can tune non-greedy ICCBF parameterizations within a unified two-stage framework, recovering fuel efficiency without sacrificing the safety certificate~\cite{wijayatunga2026twostage}. This was subsequently generalized through meta-reinforcement learning (meta-RL), in which a recurrent policy is trained over distributions of hidden physical parameters and disturbances to shape the full inner safe set online~\cite{wijayatunga2026metareinforcementlearningrobustnongreedy}. That study established that a learned, memory-based parameterization reduces conservatism and fuel consumption relative to fixed ICCBFs while preserving safety, with a recurrent Long Short-Term Memory (LSTM) policy proving especially effective in the more complex, partially observed inspection task. This paper extends that framework by systematically investigating the design space of the learned safety filter to identify the best setup for tuning the ICCBF class-$\mathcal{K}$ functions via meta-RL. It investigates: 
\begin{enumerate}
    \item \textbf{Sequence-modeling architecture.} Three recurrent network architectures including the LSTM used in prior work, the lighter Gated Recurrent Unit (GRU)~\cite{chung2014empiricalevaluationgatedrecurrent}, and the selective state-space model (Mamba)~\cite{dao2024transformersssmsgeneralizedmodels} with linear-time inference are benchmarked against one another to determine which best balances safety, fuel efficiency, and task completion within onboard computational limits, demonstrating that this choice is a decisive rather than incidental design decision.
    \item \textbf{Training algorithm.} On-policy Proximal Policy Optimization (PPO)~\cite{schulman2017proximalpolicyoptimizationalgorithms} is compared against the off-policy, entropy-regularized Soft Actor-Critic (SAC)~\cite{haarnoja2018softactorcriticoffpolicymaximum} under identical conditions to characterize the fuel--safety trade-off that each induces.
    \item \textbf{Adversarial robustness.} In addition to cooperative test cases, adversarial docking and adversarial inspection scenarios are introduced, in which the target spacecraft deliberately maneuvers to worsen the safety of the chaser or deny sensor coverage, and the robustness of each architecture--algorithm configuration is assessed.
\end{enumerate}
 
All combinations are validated through Monte Carlo studies on one-dimensional cruise control, two-dimensional docking, and three-dimensional inspection under distributions of hidden parameters, state and thrust uncertainties.  For the docking and inspection, adversarial behavior cases are also investigated.

\section{Theoretical Background}
ICCBFs are a mathematical framework for constructing input-admissible inner safe sets.  They account for actuation limits by restricting the safe state space to a smaller inner safe set. This guarantees that for every state within this inner set, there exists a feasible control command that respects physical thrust constraints while keeping the system and its future states safe. While traditional ICCBFs have fixed 
class-$\mathcal{K}$ functions, the heirachy of class-$\mathcal{K}$ functions can be learned using RL, as done in  Ref. \citenum{wijayatunga2026metareinforcementlearningrobustnongreedy}.


The system dynamics are governed by the control-affine system
\begin{equation}
\label{control_affine_system}
    \mathbf{\dot{x}} = \mathbf{f}(\mathbf{x}) + \mathbf{g}(\mathbf{x})\mathbf{u},
\end{equation}
 
\noindent where $\mathbf{f}$ and $\mathbf{g}$ are sufficiently smooth, $\mathbf{x} \in \mathcal{X} \subset \mathbb{R}^n$, and $\mathbf{u} \in \mathcal{U} \subset \mathbb{R}^m$. Here, $\mathbf{x}$ is the system state, $\mathbf{f(x)}$ the natural drift, $\mathbf{g(x)}$ the control effectiveness (the degree to which the control influences the evolution of the state), and $\mathbf{u}$ the control input.

Because the goal of ICCBFs is to keep the system safe, a safety function $h(\mathbf{x})$ is defined that maps the state to a safety score, and the effect of the environment dynamics and control inputs on that score is predicted over time. Lie derivatives describe the rate of change of a function along the system trajectory, enabling prediction of the change in the safety score under both the natural drift (when no control is applied) and the applied control. The Lie derivatives of $h(\mathbf{x})$ with respect to $\mathbf{f}(\mathbf{x})$ and $\mathbf{g}(\mathbf{x})$ are
\begin{equation}
\label{lie_derivatives}
    L_\mathbf{f} h(\mathbf{x}) = \nabla h(\mathbf{x}) \cdot \mathbf{f(x)}, \qquad L_\mathbf{g} h(\mathbf{x}) = \nabla h(\mathbf{x}) \cdot \mathbf{g(x)}.
\end{equation}
 
Using Eq.~\eqref{lie_derivatives}, the rate of change of the safety function is
\begin{equation}
    \dot{h}(\mathbf{x}) = L_\mathbf{f} h(\mathbf{x}) + L_\mathbf{g} h(\mathbf{x})\mathbf{u}.
\end{equation}

\subsection{Input-Constrained CBFs}
The safe set, represented by a continuously differentiable function $h(\mathbf{x})$, does not inherently account for physical actuation limits. Under classical Control Barrier Function (CBF) theory, forward invariance of this safe set is guaranteed if the control input $\mathbf{u}$ satisfies
\begin{equation}
    L_\mathbf{f} h(\mathbf{x}) + L_\mathbf{g} h(\mathbf{x}) \mathbf{u} \ge -\alpha(h(\mathbf{x})),
\end{equation}
 
\noindent where $\alpha$ is a class-$\mathcal{K}$ function. However, if a spacecraft lies on the boundary of the safe set while traveling with high tangential momentum, the control input required to satisfy this classical condition may exceed the physical thrust limits of the system. The dynamics would then prevent the system from remaining safe at subsequent timesteps. The ICCBF framework therefore strengthens standard CBF theory by constructing a stricter condition, $b_N(\mathbf{x})$, that prevents the system from approaching the boundary of the original safe set with excessive momentum. Beginning from the foundational safety rule $b_0(\mathbf{x}) = h(\mathbf{x})$, input awareness is introduced by computing a recursive sequence of functions that evaluates the current condition against the system dynamics and the input constraints $\mathcal{U}$, using a hierarchy of class-$\mathcal{K}$ functions $\{\alpha_i\}_{i=0}^{N-1}$:
\begin{equation}
    b_{i+1}(\mathbf{x}) = \inf_{\mathbf{u} \in \mathcal{U}} \left[ L_{\mathbf{f}} b_i(\mathbf{x}) + L_{\mathbf{g}} b_i(\mathbf{x}) \mathbf{u} + \alpha_i(b_i(\mathbf{x})) \right].
\end{equation}
 
The recursion is repeated, generating progressively stricter functions $b_1, b_2, \dots, b_N$, until it yields a subset that is intrinsically forward-invariant under the bounded inputs. This final, converged function defines the operational inner safe set, and forward invariance is enforced by any locally Lipschitz feedback satisfying
\begin{equation}
    L_\mathbf{f} b_N(\mathbf{x}) + L_\mathbf{g} b_N(\mathbf{x}) \mathbf{u} \ge -\alpha_N(b_N(\mathbf{x})).
\end{equation}


It is this learned hierarchy $\{\alpha_i\}_{i=0}^{N}$ that the meta-RL policy parameterizes.

\subsection{Time-Sampled Execution}
The system dynamics evolve in continuous time, as in Eq.~\eqref{control_affine_system}. Sensing and control updates, however, are typically executed at discrete instants on digital hardware. Intuitively, this is analogous to a person navigating a maze who can open their eyes only briefly once every few seconds: during the intervals when their eyes are closed, no guarantee can be made that they will not collide with a wall.
 
Let $t_k=kT$ denote the sampling instants, where $T>0$ is the constant sampling time step. Under a time-sampled implementation with a zero-order hold (ZOH), the control input is updated only at $t_k$ and held constant between updates, i.e, 
\begin{equation}
    \mathbf{u}(t) = \mathbf{u}_k, \quad t \in [t_k, t_{k+1}), \quad \mathbf{u}_k \in \mathcal{U}
\end{equation}
\noindent where $\mathbf{u}_k$ is computed from the sampled state $\mathbf{x}_k \triangleq \mathbf{x}(t_k)$. The resulting trajectory segment on the interval $[t_k, t_{k+1})$ is the solution to the continuous dynamics with the initial condition $\mathbf{x}(t_k)=\mathbf{x}_k$ and the constant input $\mathbf{u}_k$. Then, the state update can be written in the following integral form.
\begin{equation}
    \mathbf{x}_{k+1} = \mathbf{x}_k + \int_0^T \left( \mathbf{f}(\mathbf{x}(t_k + \tau)) + \mathbf{g}(\mathbf{x}(t_k + \tau)) \mathbf{u}_k \right) d\tau
\end{equation}
Note that enforcing a continuous-time CBF exclusively at the sampling instants does not guarantee that the system will stay in the safe set during the inter-sample period $(t_k, t_{k+1})$ unless inter-sample effects are explicitly bounded or incorporated into the condition \cite{wijayatunga2026metareinforcementlearningrobustnongreedy}. In this work, a control margin is utilized to accomplish this.

\subsection{Meta-Reinforcement Learning}
Reinforcement learning (RL) treats sequential decision-making as an agent interacting with an environment and refining its behavior through trial and error \cite{Sutton1998}. Meta-RL generalizes this from learning a single policy for a single task $\mathcal{M}$ to learning an \emph{adaptation mechanism} over a distribution of related tasks $\mathcal{M}\sim p(\boldsymbol{\mathcal{M}})$, where each $\mathcal{M}$ is drawn from a task family $\boldsymbol{\mathcal{M}}$ \cite{beck2025tutorial}. Tasks within the family differ in parameters such as mass, thrust limits, or disturbance levels, so the behavior that is optimal for a given instance depends on properties that are not fully revealed by an instantaneous observation.
 
A policy $\pi_\theta$ can take the form of either a feed-forward mapping or a recurrent dynamical system. In the feed-forward case, a multilayer perceptron (MLP) maps the current observation directly to an action, $\boldsymbol{a}_k=\pi_\theta(\boldsymbol{o}_k)$, producing a memoryless decision rule. In the recurrent case, an internal hidden state augments the mapping, $\boldsymbol{a}_k=\pi_\theta(\boldsymbol{o}_k,\boldsymbol{s}_k)$ with $\boldsymbol{s}_{k+1}=\phi_\theta(\boldsymbol{s}_k,\boldsymbol{o}_k,\boldsymbol{a}_k)$, where $\boldsymbol{s}_k$ is the hidden state and $\phi_\theta$ is its update function, allowing the policy to accumulate information across timesteps. This memory enables within-episode inference of the hidden task parameters and the corresponding adaptation, which makes recurrent policies well suited to meta-RL \cite{GAUDET2020180}.

\section{Methodology}
Given the system in Eq.~\eqref{control_affine_system}, the goal of this work is to extend the study of the meta-RL ICCBF framework in Ref. \citenum{wijayatunga2026metareinforcementlearningrobustnongreedy}. This framework consists of three parts.

\begin{enumerate}
    \item The ICCBF recursion that creates a smaller subset of the safe set
    \item A time-sampled margin, which maintains forward invariance despite discrete control updates.
    \item An agent able to output dynamic, learned class-$\mathcal{K}$ parameters, $\theta_{i,k}$, where $\alpha_i(s_k) = \theta_{i,k}$.
\end{enumerate}

A single convex quadratic program (QP) computes the control at each time step. This study considers agents that combine two RL algorithms (PPO and SAC) with three recurrent network architectures (LSTM, GRU, and Mamba), with performance measured through fuel savings and the safety score.

\subsection{Quadratic Program Formulation}
At runtime, the safe set is enforced by solving a convex QP at each sampling instant $k$. The QP modifies the nominal control command to ensure constraint satisfaction while respecting the actuation limits $\mathbf{u}_k \in \mathcal{U}$. When the controller's only active objective is to maintain safety, the baseline QP minimizes the total control effort $\|\mathbf{u}_k\|^2$ subject to the time-sampled ICCBF constraint \cite{wijayatunga2026metareinforcementlearningrobustnongreedy}, which incorporates both the learned terminal class-$\mathcal{K}$ gain $\theta_{N,k}$ and the DA-computed inter-sample margin $\hat{\nu}_k(T, \mathbf{x}_k)$, defined below.

Enforcing a continuous-time ICCBF only at the sampling instants does not guarantee safety during the inter-sample interval $(t_k, t_{k+1})$ under a zero-order hold \cite{wijayatunga2026metareinforcementlearningrobustnongreedy}. To recover strict forward invariance for all $t \ge 0$, a control margin $\hat{\nu}_k(T, \mathbf{x}_k)$ is added to the discrete ICCBF inequality, bounding the worst-case evolution of the barrier over one sample step. Computing this margin exactly requires the local Lipschitz constants of the barrier terms and the maximum bounds of the dynamics, which in turn demand numerically intensive grid-search maximization at every step \cite{Breeden_2022}. Following the formulation of Wijayatunga et al.~\cite{wijayatunga2026metareinforcementlearningrobustnongreedy}, Differential Algebra (DA) is instead used to compute conservative interval enclosures of the required Lipschitz constants and dynamics bounds over a local hyper-rectangle around $\mathbf{x}_k$, yielding an efficiently computed upper bound $\hat{\nu}_k(T, \mathbf{x}_k)$. Replacing the exact margin with this bound allows the time-sampled safety filter to be resolved with a single lightweight convex QP per control cycle. The complete derivation of this margin is given in Ref. \citenum{wijayatunga2026metareinforcementlearningrobustnongreedy}.

For goal-directed tasks, a relaxed CLF constraint is added to the QP to encode convergence to the target without compromising safety. The resulting formulation is
\begin{equation}
\begin{aligned}
    \mathbf{u}_k^* = \arg\min_{\mathbf{u}_k \in \mathbb{R}^m,\, \epsilon_k \ge 0} \quad & \tfrac{1}{2} \|\mathbf{u}_k\|^2 + p\, \epsilon_k^2 \\
    \text{s.t.} \quad & L_\mathbf{f} b_N(\mathbf{x}_k) + L_\mathbf{g} b_N(\mathbf{x}_k)\mathbf{u}_k + \theta_{N,k}\, b_N(\mathbf{x}_k) \ge \hat{\nu}_k(T, \mathbf{x}_k), \\
    & L_\mathbf{f} V(\mathbf{x}_k) + L_\mathbf{g} V(\mathbf{x}_k)\mathbf{u}_k \le -c_{V,k} V(\mathbf{x}_k) + \epsilon_k, \\
    & \mathbf{u}_k \in \mathcal{U},
\end{aligned}
\end{equation}
 
\noindent where $p > 0$ weights the CLF slack variable $\epsilon_k$. Because the terminal ICCBF condition is affine in the control input, the optimization is a strictly convex QP, ensuring the computational efficiency required for high-rate flight execution \cite{wijayatunga2026metareinforcementlearningrobustnongreedy}.

\subsection{Meta-RL Formulation}
The parameter-tuning problem is posed as an episodic, discrete-time control task. At each sampling instant $t_k$, the agent observes the system state and outputs the ICCBF class-$\mathcal{K}$ parameters $\boldsymbol{\theta}_k$ (and, where a CLF is present, the gain $c_{V,k}$), which together shape the geometry of the inner safe set enforced by the QP.
 
\subsubsection{Network Architecture and Sequence Modeling}
Because the optimal class-$\mathcal{K}$ parameters depend on physical properties that are not observable from a single state, the policy must integrate the trajectory history to infer the hidden dynamics. Both the actor and critic therefore use separate feature extractors followed by an independent recurrent sequence module. To assess how the choice of temporal-inference mechanism affects performance, independent agents are trained with three sequence-modeling network architectures: an LSTM, a GRU, and a Mamba (selective state-space) module. For each scenario, one policy is trained per backbone with each of the two algorithms (PPO and SAC), giving six training runs per scenario. The PPO and SAC update procedures are detailed in Appendix~A.
 
\subsubsection{Domain Randomization and Uncertainty}
To train a policy that generalizes across the task family, each episode randomizes the initial state, the hidden physical parameters, and the sensing and actuation noise. These distributions are detailed below.
 
\subsubsection{Initial State and Hidden Parameter Distributions}
Each training episode begins by drawing an initial condition together with a set of environmental parameters, such as the deputy mass, the maximum thrust $u_{\max}$, and the keep-in/keep-out zone sizes, so that the meta-RL agent is exposed to a distribution of task instances rather than a single nominal scenario, which improves robustness and limits overfitting. The initial state $\mathbf{x}_0$ is drawn from a bounded admissible set $\mathcal{X}_0 \subseteq \mathcal{S}$, and a vector of hidden parameters $\boldsymbol{p} \in \mathbb{R}^{n_p}$ is sampled from a hyper-rectangle centered on the nominal values $\boldsymbol{\bar p}$. Each component is sampled independently as
\begin{equation}\label{eq:drawparam}
p_i \sim \mathrm{Uniform}\!\left(\bigl[(1-\delta_i)\,\bar p_i,\ (1+\delta_i)\,\bar p_i\bigr]\right), \quad i = 1,\dots,n_p,
\end{equation}
 
\noindent where $\bar p_i$ is the nominal value of the $i$th parameter and $\delta_i > 0$ sets its relative variation. The per-case values of $\delta_i$ are listed in Table~\ref{tab:uncertainty_summary}. Within each Monte Carlo dataset, the parameter samples are held fixed across all controllers, so that performance differences reflect only the ICCBF tuning mechanism.
 
\begin{table}[t]
\centering
\small
\caption{Hidden parameter variations per episode (nominal value with relative deviation $\delta_i$). Adapted from Ref.~\cite{wijayatunga2026metareinforcementlearningrobustnongreedy}.}
\label{tab:uncertainty_summary}
\begin{tabular}{lccc}
\toprule
\textbf{Category} & \textbf{Cruise Control} & \textbf{Docking} & \textbf{Inspection} \\
\midrule
Deputy mass $m$            & $1650~\mathrm{kg}$ ($\pm 20\%$)        & $1000~\mathrm{kg}$ ($\pm 10\%$) & $12~\mathrm{kg}$ ($\pm 10\%$) \\
Max thrust $u_{\max}$       & $0.25~\mathrm{N}$ ($\pm 20\%$)         & $0.25~\mathrm{kN}$ ($\pm 10\%$) & $1.0~\mathrm{N}$ ($\pm 10\%$) \\
Max velocity $v_{\max}$     & $24.0~\mathrm{m\,s^{-1}}$ ($\pm 20\%$) & --                              & -- \\
Lead vehicle velocity $v_0$ & $13.89~\mathrm{m\,s^{-1}}$ ($\pm 10\%$)& --                              & -- \\
Deputy radius $R_D$         & --                                     & --                              & $5~\mathrm{m}$ ($\pm 10\%$) \\
Chief radius $R_C$          & --                                     & $2.4~\mathrm{m}$ ($\pm 10\%$)   & $10~\mathrm{m}$ ($\pm 10\%$) \\
Chief spin rate $\omega$    & --                                     & $0.6^\circ\,\mathrm{s^{-1}}$ ($\pm 10\%$) & -- \\
Chief orbit radius $r$      & --                                     & $6771~\mathrm{km}$ ($\pm 10\%$) & $6771~\mathrm{km}$ ($\pm 10\%$) \\
Approach cone half-angle $\gamma$ & --                               & $10^\circ$ ($\pm 10\%$)         & -- \\
KIZ radius $R_{\max}$       & --                                     & --                              & $800~\mathrm{m}$ ($\pm 10\%$) \\
\bottomrule
\end{tabular}
\end{table}
 
\subsubsection{State Errors}
Sensing errors are modeled as additive zero-mean Gaussian noise on the true state. At each step $k$, the state presented to the agent is
\begin{equation}\label{eq:statenoise}
    \boldsymbol{x}_k^E = \boldsymbol{x}_k + \boldsymbol{\epsilon}_{\boldsymbol{x}_k},
    \qquad
    \boldsymbol{\epsilon}_{\boldsymbol{x}_k} \sim \mathcal{N}\!\left(\boldsymbol{0},\, \boldsymbol{\sigma}_x^2 \boldsymbol{I}\right),
\end{equation}
 
\noindent where $\boldsymbol{\sigma}_x = [\boldsymbol{\sigma}_r, \boldsymbol{\sigma}_v]^\top$ collects the position- and velocity-error standard deviations listed in Table~\ref{tab:noise_models}.
 
\subsubsection{Thrust Errors}
Actuation errors are introduced by perturbing the magnitude and direction of the commanded control $\boldsymbol{u}_k^\star$ before propagation. With $\boldsymbol{u}_k^\star = [u_x, u_y, u_z]^\top$ expressed in the control frame, the magnitude $u_k = \|\boldsymbol{u}_k^\star\|_2$ and the out-of-plane and in-plane angles $(\beta, \gamma)$ are perturbed independently, and the executed command is reconstructed and saturated to $u_{\max}$:
\begin{align}\label{eq:execution_noise_model}
u_k &= \|\boldsymbol{u}_k^\star\|_2,\qquad
\beta = \sin^{-1}\!\left(\frac{u_z}{u_k}\right),\qquad
\gamma = \tan^{-1}\!\left(\frac{u_x}{u_y}\right), \\
u_{k}^{E} &= u_{k}\left(1+\delta_u\right),\qquad
\beta^{E} = \beta + \delta_{\beta},\qquad
\gamma^{E} = \gamma + \delta_{\gamma}, \\
\boldsymbol{u}_k &= u_{k}^{E}
\begin{bmatrix}
\cos \beta^{E}\,\sin \gamma^{E}\\
\cos \beta^{E}\,\cos \gamma^{E}\\
\sin \beta^{E}
\end{bmatrix},
\qquad
\boldsymbol{u}_k \leftarrow \frac{u_{\max}}{\max(u_{\max},\|\boldsymbol{u}_k\|_2)}\,\boldsymbol{u}_k,
\end{align}
 
\noindent where $\delta_u \sim \mathcal{N}(0,\sigma_u^2)$, $\delta_{\beta} \sim \mathcal{N}(0,\sigma_\beta^2)$, and $\delta_{\gamma} \sim \mathcal{N}(0,\sigma_\gamma^2)$. The corresponding standard deviations are given in Table~\ref{tab:noise_models}.
 
\begin{table}[t]
\centering
\caption{Standard deviations of the thrust and state uncertainties applied.}
\label{tab:noise_models}
\begin{tabular}{lccc}
\toprule
\textbf{Noise source} & \textbf{Cruise Control} & \textbf{Docking} & \textbf{Inspection} \\
\midrule
$\sigma_u$              & $0.1$                   & $0.05$            & $0.05$ \\
$\sigma_\gamma$         & --                      & $0.1^\circ$       & $0.1^\circ$ \\
$\sigma_\beta$          & --                      & --                & $0.1^\circ$ \\
$\sigma_r$ (each axis)  & $2~\mathrm{m}$          & $0.1~\mathrm{m}$  & $0.1~\mathrm{m}$ \\
$\sigma_v$ (each axis)  & $0.5~\mathrm{m\,s^{-1}}$ & $2~\mathrm{mm\,s^{-1}}$ & $2~\mathrm{mm\,s^{-1}}$ \\
\bottomrule
\end{tabular}
\end{table}
 
\subsubsection{Observation}
At each step, the environment supplies an observation $\boldsymbol{S}_k$ carrying enough information to estimate the value of a candidate action. For numerical stability and consistent feature scaling, the observation is min--max normalized to the interval $[-1, 1]$:
\begin{equation}\label{eq:scaleobs}
\boldsymbol{S}_k^* = 2\!\left(\frac{\boldsymbol{S}_k - \boldsymbol{S}_{\min}}{\boldsymbol{S}_{\max} - \boldsymbol{S}_{\min}}\right) - 1,
\end{equation}
 
\noindent where $\boldsymbol{S}_{\min}$ and $\boldsymbol{S}_{\max}$ bound the components of $\boldsymbol{S}$. For the cruise-control and docking cases, the observation is the noisy state, $\boldsymbol{S}_k = \boldsymbol{x}_k^E$. For inspection it is augmented to $\boldsymbol{S}_k = [\boldsymbol{x}_k^E,\ \theta_S,\ n_{\text{insp}},\ \hat{\boldsymbol{d}}]^\top$, where $\theta_S$ is the Sun angle, $n_{\text{insp}}$ the number of inspected points, and $\hat{\boldsymbol{d}}$ a unit vector toward the largest cluster of remaining uninspected points, obtained by K-means clustering \cite{doi:10.2514/1.I011391}.
 
\subsubsection{Reward Function}
The agent's objective is to minimize control effort (fuel) while strictly avoiding task failure and maintaining safety \cite{wijayatunga2026metareinforcementlearningrobustnongreedy}. The per-step reward penalizes the magnitude of the commanded nominal control $\|\mathbf{u}_k^*\|_2$, promoting fuel-efficient trajectories, together with a failure flag and a safety-violation term:
\begin{equation}
    \label{eq:per_step_reward}
    r_k = -w_u \|\mathbf{u}_k^*\|_2 - w_{\text{fail}} P_{\text{fail}} - w_h \max\!\left(0,\, -h_{k+1}\right),
\end{equation}
 
\noindent where $w_u$, $w_{\text{fail}}$, and $w_h$ are positive weights, $h_{k+1}$ is the safety value at the next state, and the failure flag terminates the episode when the QP solver becomes infeasible or a safety constraint is violated,
\begin{equation}
    \label{eq:fail_flag}
    P_{\text{fail}} =
    \begin{cases}
        1, & \text{QP infeasible or solver failure}, \\
        0, & \text{otherwise}.
    \end{cases}
\end{equation}
 
For tasks that use a CLF to encode goal-reaching, a terminal penalty is added at the episode horizon $t_f$ to penalize failure to converge to the target:
\begin{equation}
    \label{eq:total_reward}
    R_k =
    \begin{cases}
        r_k, & t_k < t_f, \\
        r_k - w_V P_V, & t_k \ge t_f,
    \end{cases}
\end{equation}
 
\noindent where $w_V > 0$ and the terminal penalty activates only if the minimum CLF value over the episode, $\min_j V(\mathbf{x}_j)$, fails to fall below a convergence threshold $\rho_V$:
\begin{equation}
    \label{eq:clf_penalty}
    P_V =
    \begin{cases}
        \min_j V(\mathbf{x}_j), & \min_j V(\mathbf{x}_j) > \rho_V, \\
        0, & \text{otherwise}.
    \end{cases}
\end{equation}
 
For the inspection scenario, which has no CLF, the per-step reward in Eq.~\eqref{eq:per_step_reward} is augmented with a positive reward proportional to the number of newly inspected surface points, encouraging coverage alongside fuel economy.

\section{Results}

\subsection{Test Cases}

\subsubsection{Cooperative Scenarios}
The three cooperative test cases including a one-dimensional cruise-control problem, a two-dimensional docking problem, and a three-dimensional inspection problem are adopted directly from the meta-RL ICCBF study of Wijayatunga et al.~\cite{wijayatunga2026metareinforcementlearningrobustnongreedy}, which in turn draws the cruise-control and docking problems from \cite{Agrawal_2021} and the inspection problem from \cite{doi:10.2514/1.I011391}. They are summarized below for completeness; the full dynamics, safety constraints, and CLFs are given in \cite{wijayatunga2026metareinforcementlearningrobustnongreedy}.
 \begin{itemize}
     \item \textbf{Cruise Control} A didactic, non-RPO benchmark in which a deputy follows a lead vehicle moving at a constant speed $v_0$. With state $\mathbf{x} = [d,\ v]^\top$ comprising the headway distance $d$ and the deputy velocity $v$, the dynamics are
\begin{equation}
\begin{bmatrix} \dot{d} \\ \dot{v} \end{bmatrix} = \begin{bmatrix} v_0 - v \\ -\tfrac{F(v)}{m} \end{bmatrix} + \begin{bmatrix} 0 \\ g_0 \end{bmatrix} u, \qquad \mathcal{U} = \{u : |u| \le u_{\max}\},
\end{equation}
 
\noindent where $F(v) = 0.1 + 5v + 0.25v^2$ is a resistive drag, $m$ is the deputy mass, and $g_0$ is a control-scaling constant. Collision avoidance defines the safe set through the CBF
\begin{equation}
h(\mathbf{x}) = d - 1.8\,v \ge 0,
\end{equation}
 
\noindent and a CLF drives the deputy toward the speed limit $v_{\max}$:
\begin{equation}
V(\mathbf{x}) = (v - v_{\max})^2.
\end{equation}
\item \textbf{Docking with a Coorperative Target.} A planar rendezvous in which a deputy must reach the rotating docking port of a chief while remaining within a line-of-sight cone (Figure~\ref{fig:docking_problem}). The state is $\mathbf{x} = [p_x,\ p_y,\ v_x,\ v_y,\ \phi]^\top$, where $(p_x, p_y)$ and $(v_x, v_y)$ are the chaser's relative position and velocity and $\phi$ is the docking-port orientation, evolving as $\dot{\phi} = \omega$. The relative translational motion follows the Clohessy--Wiltshire (CW) dynamics with thrust acceleration as the control input,
\begin{equation}
\begin{aligned}
&\dot{p}_x = v_x, \qquad \dot{p}_y = v_y, \qquad \dot{\phi} = \omega, \\
&\dot{v}_x = 3n^2 p_x + 2n v_y + \tfrac{u_x}{m}, \qquad \dot{v}_y = -2n v_x + \tfrac{u_y}{m},
\end{aligned}
\end{equation}
 
\noindent where $n$ is the chief's mean motion and $\|\mathbf{u}\| \le u_{\max}$. The line-of-sight requirement defines the CBF
\begin{equation}
h(\mathbf{x}) = \frac{\bar{\mathbf{r}}_{cp} \cdot \hat{\mathbf{e}}}{\|\bar{\mathbf{r}}_{cp}\|} - \cos\gamma \ge 0,
\end{equation}
 
\noindent with cone half-angle $\gamma$, cone axis $\hat{\mathbf{e}} = [\cos\phi,\ \sin\phi]^\top$, port offset $\rho$, and $\bar{\mathbf{r}}_{cp} = [p_x - \rho\cos\phi,\ p_y - \rho\sin\phi]^\top$ the vector from the port to the chaser. Convergence to the port is encoded by the CLF
\begin{equation}
V(\mathbf{x}) = \left(v_x + \frac{p_x - \rho\cos\phi}{10}\right)^2 + \left(v_y + \frac{p_y - \rho\sin\phi}{10}\right)^2.
\end{equation}
 
\begin{figure}[hbt]
    \centering
    \includegraphics[width=0.7\linewidth]{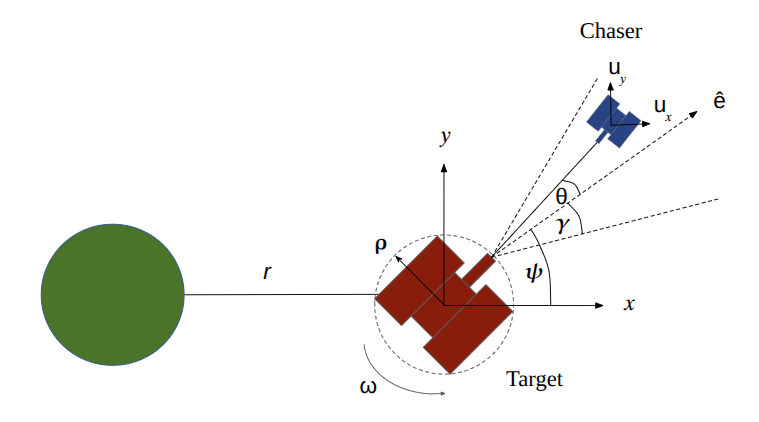}
    \caption{Spacecraft docking problem.}
    \label{fig:docking_problem}
\end{figure}
 \item \textbf{Inspecting a Cooperative Target.} A three-dimensional problem in which a deputy circumnavigates a spherical chief to inspect discretized points on its surface (Figure~\ref{fig:inspection_scenario}). With state $\mathbf{x} = [\mathbf{r}^\top,\ \mathbf{v}^\top]^\top \in \mathbb{R}^6$, where $\mathbf{r} = [x,\ y,\ z]^\top$ and $\mathbf{v}$ are the relative position and velocity in the chief-centered LVLH frame, the motion follows the CW equations
\begin{equation}
\ddot{x} = 3n^2 x + 2n\dot{y} + \tfrac{u_x}{m}, \qquad \ddot{y} = -2n\dot{x} + \tfrac{u_y}{m}, \qquad \ddot{z} = -n^2 z + \tfrac{u_z}{m},
\end{equation}
 
\noindent with $\|\mathbf{u}\|_\infty \le u_{\max}$. Three constraints define the safe set---a keep-out zone (KOZ), a keep-in zone (KIZ), and a sensor Sun-exclusion condition:
\begin{align}
h_1(\mathbf{x}) &= \|\mathbf{r}\| - (R_C + R_D) \ge 0, \\
h_2(\mathbf{x}) &= R_{\max} - \|\mathbf{r}\| \ge 0, \\
h_3(\mathbf{x}) &= \theta_b - \alpha_{\mathrm{FOV}}/2 \ge 0,
\end{align}
 
\noindent where $R_C$ and $R_D$ are the chief and deputy radii, $R_{\max}$ is the KIZ radius, $\theta_b$ is the sensor boresight--Sun angle, and $\alpha_{\mathrm{FOV}}$ is the sensor field of view. Because surface coverage cannot be expressed as a CLF, a learned nominal control policy replaces the CLF term and the ICCBF acts as a supervisory safety filter.
 
\begin{figure}[hbt]
    \centering
    \includegraphics[width=0.7\linewidth]{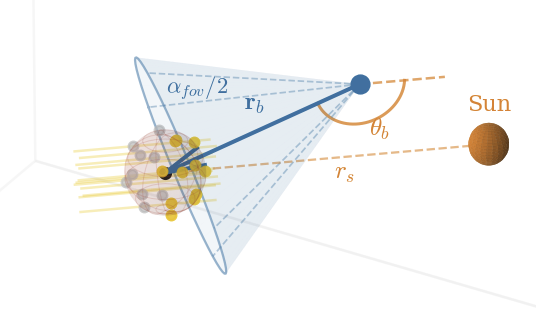}
    \caption{Spacecraft inspection problem. $r_s$ and $r_b$ denote the vector that points from the Sun to the target and the vector that points from the chaser to the target respectively. $\alpha_{FOV}$ denotes the field of view of the chaser, and $\theta_b$ is the angle between the chaser's sensor boresight and the Sun.}
    \label{fig:inspection_scenario}
\end{figure}

 \end{itemize}

\subsubsection{Adversarial Scenarios}
In addition to cooperative docking and inspection scenarios, in which the target remains passive, this section explores the same test cases under adversarial target behavior.

\begin{itemize}
    \item \textbf{Docking with an Adversarial Target.} The target acts as an adversary by choosing $\omega_{k+1} = \dot{\phi}_{k+1}$ to minimize $h(\boldsymbol x)$. After the $k$th step, it evaluates $h$ under a small perturbation $\Delta\omega$ to the current rotation rate.
\begin{align}
    h_0^+ &= h_0\!\left(p_{x,k+1},\, p_{y,k+1},\, v_{x,k+1},\, v_{y,k+1},\, \phi_{k+1}+(\omega_k+\Delta\omega)T\right) \\
    h_0^- &= h_0\!\left(p_{x,k+1},\, p_{y,k+1},\, v_{x,k+1},\, v_{y,k+1},\, \phi_{k+1}+(\omega_k-\Delta\omega)T\right)
\end{align}
\noindent where $T$ is the simulation timestep. The target then applies an update to its rotation rate in the direction that reduces safety as follows
\begin{equation}
    \omega_{k+1}=
    \begin{cases}
      \omega_k - \Delta\omega_{\max} & \text{if } h_0^- < h_0^+ ,\\
      \omega_k + \Delta\omega_{\max} & \text{if } h_0^- \geq h_0^+.
    \end{cases}
\end{equation}
Then, it clips to admissible bounds such that 
\begin{equation}
    \omega_{k+1} = \text{clip}\!\left(\omega_{k+1},\, \omega_{\min},\, \omega_{\max}\right).
\end{equation}
We use $[\omega_{\min}, \omega_{\max}] = [0,\ 0.7\ \text{deg/s}]$ and $\Delta\omega = \Delta\omega_{\max} = 0.02\ \text{deg/s}$.
The chaser's ICCBF controller only receives a the stale estimate of $\omega_k$ that lags by one time-step.
\item \textbf{Inspecting an Adversarial Target.} The chief applies a small impulsive maneuver at each step that increases the relative separation, modeled as a disturbance on the deputy's discrete CW dynamics directed radially outward:
\begin{equation}
    \mathbf{x}_{k+1} = \Phi(T)\mathbf{x}_k + \Gamma(T)\mathbf{u}_{\text{dep},k} - \Gamma(T)\mathbf{d}_k,
    \qquad
    \boldsymbol{d}_k = -\Delta v_k \, \frac{\mathbf{r}_k}{\|\mathbf{r}_k\|},
\end{equation}
 
\noindent where $\Phi(T)$ and $\Gamma(T)$ are the CW state-transition and input matrices, $\mathbf{u}_{\text{dep},k}$ is the deputy control, and $\mathbf{r}_k = [p_x, p_y, p_z]^\top$ is the deputy position relative to the chief, so that $\mathbf{d}_k$ pushes the deputy away from the target. At episode reset, a total maneuver budget $\Delta v_{\text{total}} \sim \mathcal{U}(0.5,\ 5.0)~\text{m/s}$ is sampled and spread evenly across the $N$-step horizon, $\Delta v_k = \Delta v_{\text{total}}/N$.

\end{itemize}

\subsection{Experimental Setup}
Training was conducted over a maximum of $100{,}000$ episodes per run, distributed across 28 to 32 parallel environments, accumulating approximately 20 million timesteps. The recurrent network architectures were trained with Backpropagation Through Time (BPTT). To mitigate recurrent-state initialization error during updates, a burn-in period (20 to 100 steps, depending on the domain) was used to warm up the hidden states before gradients were accumulated. The primary hyperparameters are summarized below.
\begin{itemize}
    \item \textbf{PPO:} learning rate $\approx 3 \times 10^{-4}$ (constant or decaying by domain), discount factor $\gamma = 0.995$, GAE $\lambda = 0.95$, clip parameter $\epsilon = 0.2$, and entropy coefficient $c_{\text{ent}} = 0.01$.
    \item \textbf{SAC:} learning rate $\approx 3 \times 10^{-4}$, target smoothing coefficient $\tau = 0.005$, automatic target-entropy tuning (floor $0.05$), and a replay buffer of $50{,}000$ to $100{,}000$ transitions.
    \item \textbf{Networks:} the MLP feature extractors used 2 to 3 hidden layers with 64 to 256 nodes per layer. For a fair comparison, the hidden-state dimension ($d_{\text{model}}$) of the LSTM, GRU, and Mamba2 modules was matched (32 to 256 units, depending on domain complexity).
\end{itemize}

\subsection{Evaluation Methodology}
\begin{figure}[hbt]
    \centering
    \begin{minipage}[t]{0.46\linewidth}
        \centering
        \includegraphics[width=\linewidth]{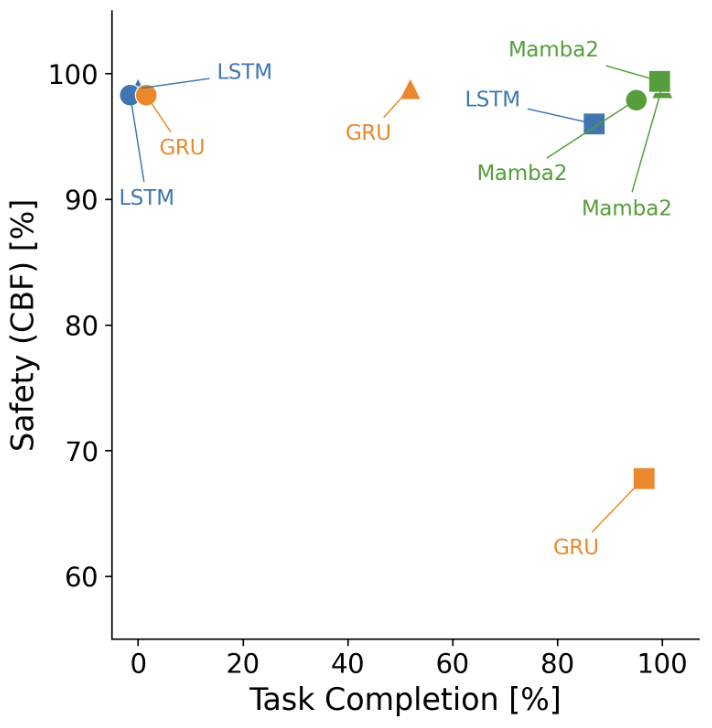}
        \caption{Performance of all setups across the three cooperative mission scenarios. Triangle: Cruise Control, Circle: Docking, Square: Inspection.}
        \label{fig:meta_scatter}
    \end{minipage}%
    \hfill
    \begin{minipage}[t]{0.46\linewidth}
        \centering
        \includegraphics[width=\linewidth]{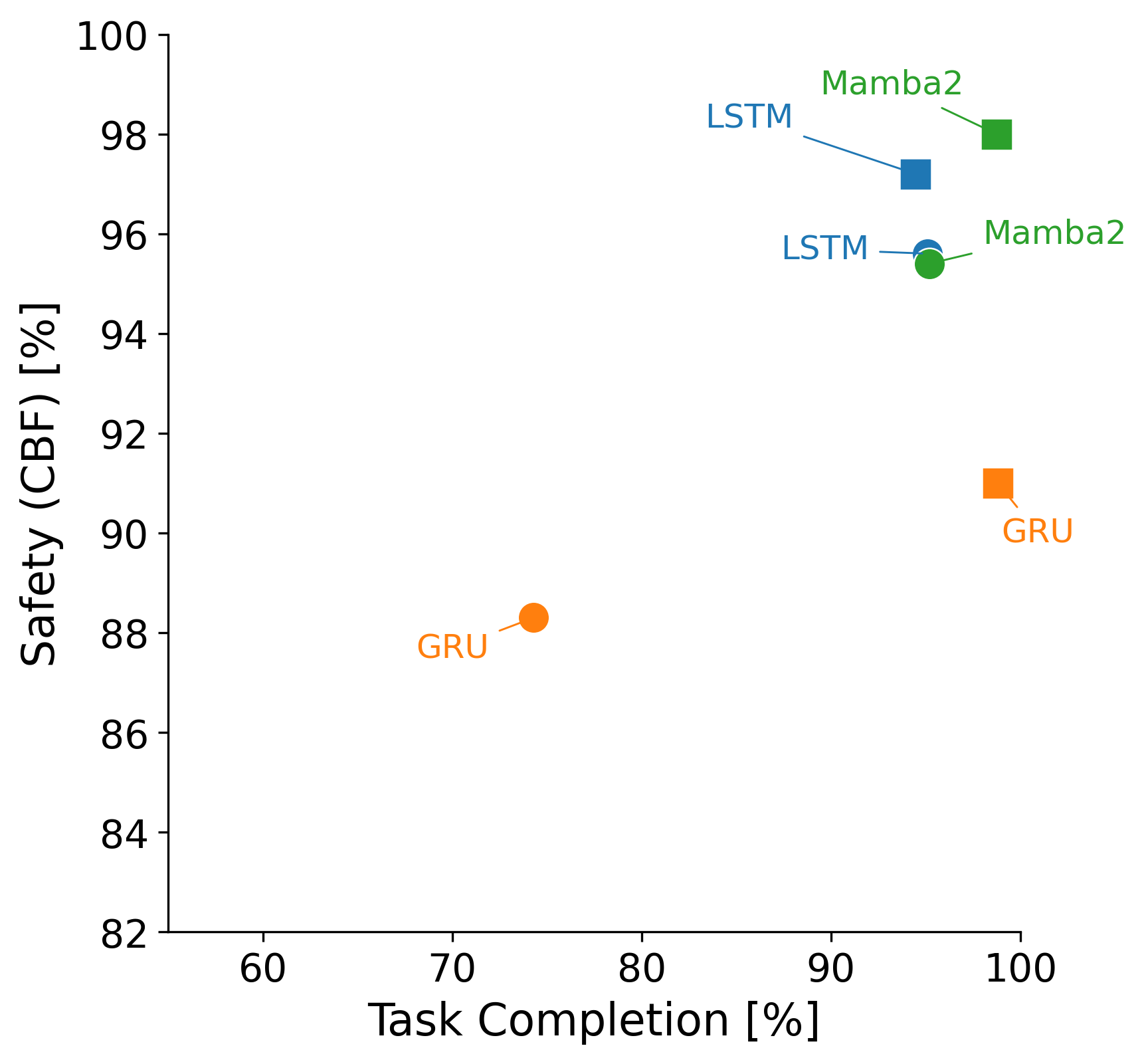}
        \caption{Performance of all setups across the two adversarial mission scenarios. Circle: Docking, Square: Inspection.}
        \label{fig:meta_scatter_adversarial}
    \end{minipage}
\end{figure}

To assess performance, safety, and efficiency, the evaluation phase was decoupled from training. For each scenario, a fixed bank of 500 pre-generated initial conditions and environment-parameter randomizations (e.g., mass and inertia) was established. Using a static bank guarantees that every policy is evaluated on the exact same sequence of test cases regardless of training algorithm or recurrent backbone, enabling a direct comparison. During evaluation, the policies were executed deterministically: for PPO, the action-distribution means were used; for SAC, the exploration noise was disabled. This ensures that performance differences arise solely from the learned representations.
Each controller was assessed using the following three metrics.
\begin{enumerate}
    \item \textbf{Safety:} the fraction of episodes in which the state remained within the safe set $h(\mathbf{x},t) \ge 0$ for the entire horizon (or until the goal was reached).
    \item \textbf{Fuel / control effort:} the accumulated control effort over the trajectory, reported as total $\Delta v$ or the integral of the control magnitude $\int \|\mathbf{u}(t)\|\,dt$.
    \item \textbf{Task performance:} domain-specific success criteria, including time of flight (TOF) to the target state, docking-angle error, and the final percentage of the target surface inspected.
\end{enumerate}
Figures~\ref{fig:meta_scatter} and \ref{fig:meta_scatter_adversarial} and Table \ref{tab:coop_uncoop_summary} summarize the results across the three cooperative scenarios and the two adversarial scenarios, respectively. Detailed results follow.
 
\begin{table*}[hbt]
\centering
\footnotesize
\setlength{\tabcolsep}{5pt}
\renewcommand{\arraystretch}{1.15}
\caption{Performance summary across the cooperative and adversarial scenarios. Fuel is reported as mean\,$\pm$\,std with quartiles $[Q_1,Q_2,Q_3]$ (N$\cdot$s for Cruise Control, m/s otherwise). ``Safe'' is the fraction of safe episodes; ``Compl.'' is the docking-success rate (Docking) or mean surface coverage (Inspection). Best value per column within each task in bold.}
\label{tab:coop_uncoop_summary}
\begin{tabular}{@{}l l r c c c@{}}
\hline
\textbf{Task} & \textbf{Case} & $\boldsymbol{\mu \pm \sigma}$ & $\boldsymbol{[Q_1,\,Q_2,\,Q_3]}$ & \textbf{Safe (\%)} & \textbf{Compl.\ (\%)} \\
\hline
\multirow{6}{*}{Cruise Control}
& LSTM+PPO   & $4.02 \pm 0.440$ & $[3.99,\,4.07,\,4.14]$ & 98.8 & -- \\
& LSTM+SAC   & $3.53 \pm 0.497$ & $[3.29,\,3.58,\,3.84]$ & 98.9 & -- \\
& GRU+PPO    & $3.29 \pm 0.514$ & $[3.04,\,3.38,\,3.63]$ & 98.8 & -- \\
& GRU+SAC    & $3.98 \pm 0.416$ & $[3.93,\,4.03,\,4.12]$ & $\mathbf{99.0}$ & -- \\
& Mamba2+PPO & $\mathbf{2.62 \pm 0.355}$ & $[\mathbf{2.47},\,\mathbf{2.60},\,\mathbf{2.79}]$ & 98.9 & -- \\
& Mamba2+SAC & $3.90 \pm 0.449$ & $[3.81,\,3.97,\,4.09]$ & 98.8 & -- \\
\hline
\multirow{6}{*}{Cooperative Docking}
& LSTM+PPO   & $7.64 \pm 2.97$ & $[5.02,\,7.15,\,10.6]$ & $\mathbf{98.3}$ & 0.00 \\
& LSTM+SAC   & $\mathbf{5.23 \pm 1.05}$ & $[\mathbf{4.53},\,\mathbf{5.09},\,\mathbf{5.97}]$ & 98.2 & 0.00 \\
& GRU+PPO    & $6.90 \pm 2.47$ & $[4.90,\,6.25,\,8.81]$ & $\mathbf{98.3}$ & 0.00 \\
& GRU+SAC    & $7.39 \pm 1.63$ & $[6.67,\,7.08,\,7.85]$ & 4.80 & 2.20 \\
& Mamba2+PPO & $6.26 \pm 1.70$ & $[5.14,\,6.02,\,7.17]$ & 97.9 & $\mathbf{95.0}$ \\
& Mamba2+SAC & $6.06 \pm 1.88$ & $[4.74,\,5.21,\,7.73]$ & $\mathbf{98.3}$ & 48.6 \\
\hline
\multirow{6}{*}{Cooperative Inspection}
& LSTM+PPO   & $8.31 \pm 4.30$   & $[6.02,\,7.04,\,8.80]$   & 94.4 & 97.7 \\
& LSTM+SAC   & $20.1 \pm 11.0$   & $[12.5,\,16.6,\,22.4]$   & 6.40 & 76.5 \\
& GRU+PPO    & $9.69 \pm 4.24$   & $[6.50,\,7.69,\,13.9]$   & 67.8 & 98.1 \\
& GRU+SAC    & $20.7 \pm 21.9$   & $[8.48,\,9.64,\,25.2]$   & 0.00 & 37.4 \\
& Mamba2+PPO & $\mathbf{6.59 \pm 1.38}$ & $[\mathbf{5.71},\,\mathbf{6.55},\,\mathbf{7.58}]$ & $\mathbf{99.4}$ & $\mathbf{99.6}$ \\
& Mamba2+SAC & $9.20 \pm 6.78$   & $[7.78,\,8.40,\,9.06]$   & 0.00 & 27.7 \\
\hline
\multirow{3}{*}{Adversarial Docking}
& LSTM+PPO   & $6.77 \pm 2.92$ & $[4.58,\,6.27,\,8.27]$ & $\mathbf{95.6}$ & 95.1 \\
& GRU+PPO    & $7.24 \pm 2.68$ & $[5.24,\,7.05,\,9.74]$ & 88.3 & 74.3 \\
& Mamba2+PPO & $\mathbf{5.18 \pm 1.87}$ & $[\mathbf{3.47},\,\mathbf{5.39},\,\mathbf{6.82}]$ & 95.4 & $\mathbf{95.2}$ \\
\hline
\multirow{3}{*}{Adversarial Inspection}
& LSTM+PPO   & $7.79 \pm 2.80$ & $[6.17,\,7.09,\,8.38]$ & 95.6 & 98.4 \\
& GRU+PPO    & $9.69 \pm 4.24$ & $[6.50,\,7.69,\,13.9]$ & 67.8 & 98.1 \\
& Mamba2+PPO & $\mathbf{5.29 \pm 1.37}$ & $[\mathbf{4.32},\,\mathbf{5.22},\,\mathbf{6.07}]$ & $\mathbf{99.0}$ & $\mathbf{99.4}$ \\
\hline
\end{tabular}
\end{table*}

\subsection{Cruise Control}
\begin{figure}[hbt]
    \centering
    \includegraphics[width=0.9\linewidth]{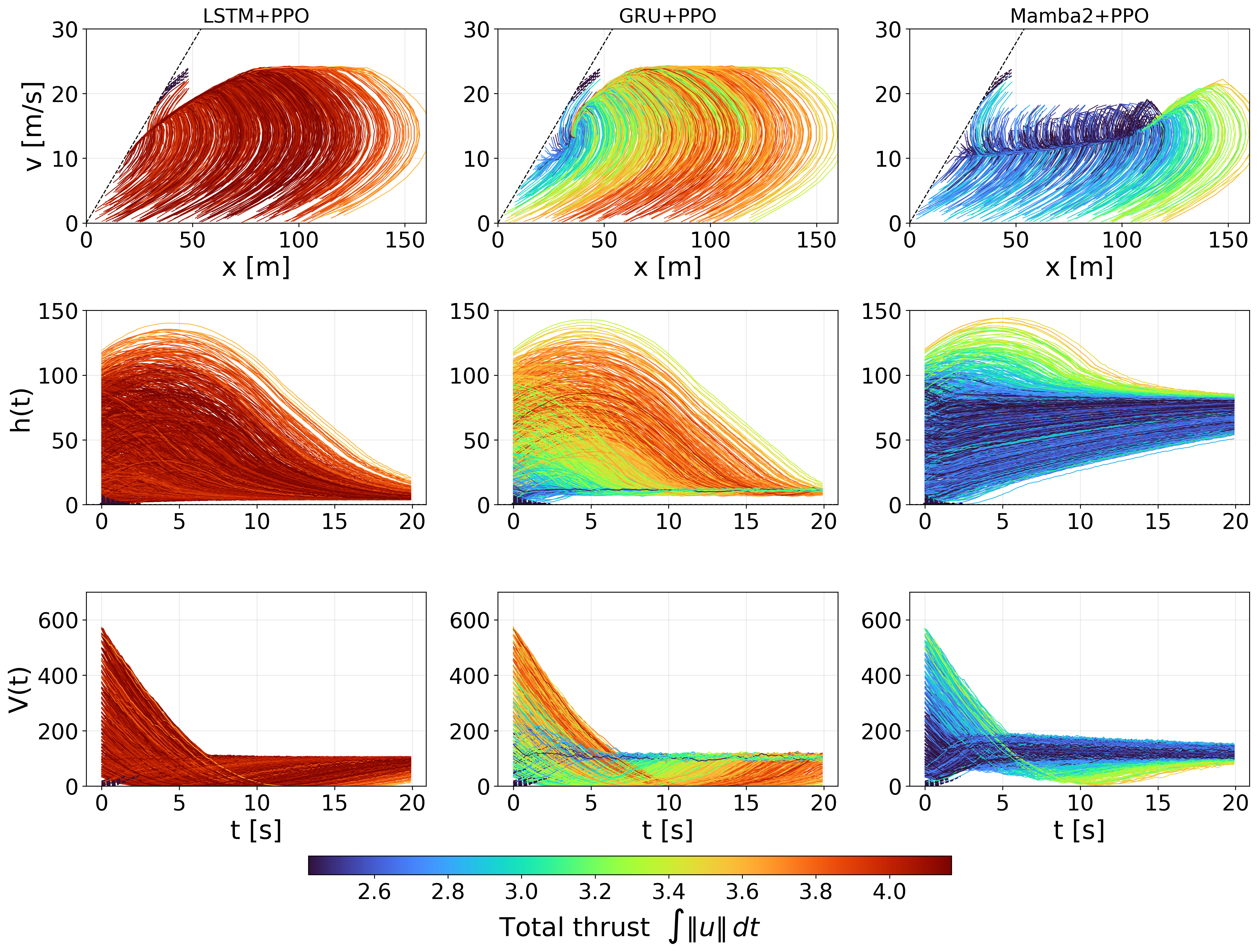}
    \caption{PPO cruise control trajectories}
    \label{fig:cc_ppo_traj}
\end{figure}

\begin{figure}[hbt]
    \centering
    \includegraphics[width=0.9\linewidth]{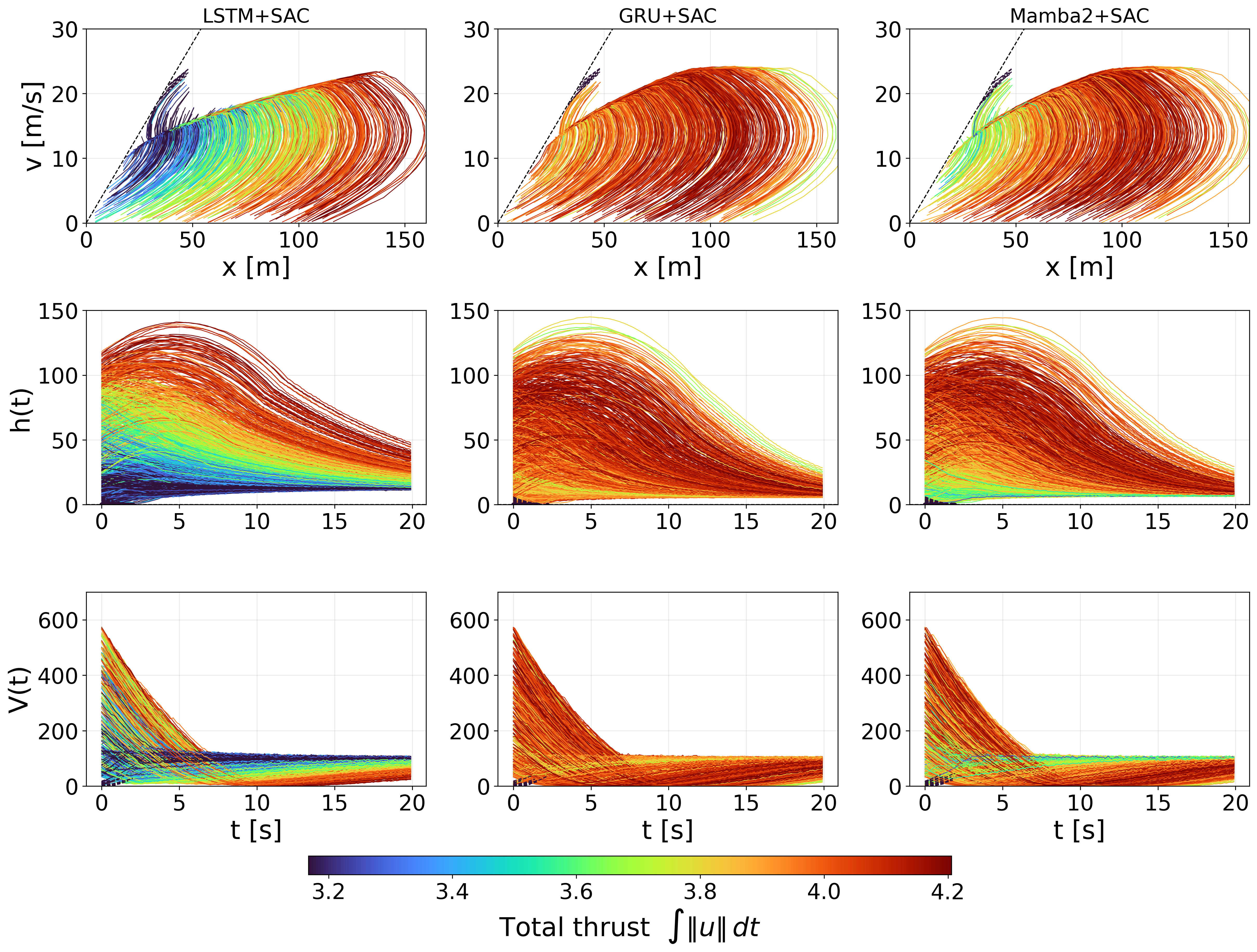}
    \caption{SAC cruise control trajectories}
    \label{fig:cc_sac_traj}
\end{figure}

As shown in Table \ref{tab:coop_uncoop_summary}, the Cruise Control task is well within the capabilities of all evaluated models. Every combination achieves a safety rate of nearly $99.0\%$. However, Mamba2+PPO distinguishes itself by exhibiting the lowest mean fuel usage. Figure~\ref{fig:cc_ppo_traj} shows the PPO-trained results, with phase-space trajectories in $(d,v)$ colored by total thrust (top), the CBF value $h(t)$ against the $h=0$ safety boundary (middle), and the Lyapunov function $V(t)$ (bottom). This figure shows that the Mamba2+PPO undercuts LSTM+PPO and GRU+PPO on fuel. All architectures fail to fully satisfy the CLF with $V(t)$ plateauing near $100$--$150$ rather than converging to zero.This behavior is a consequence of the reward structure prioritizing fuel over the CLF leading the agent to reduce its velocity during its approach. In future work, higher weighting for the CLF satisfaction condition in the RL reward will be explored to avoid this.  Figure~\ref{fig:cc_sac_traj} shows the analogous results for the three SAC-trained network architectures, but the median fuel consumptions are considerably higher than PPO's.



\subsection{Docking with a Coorperative Target}
\begin{figure}[hbt]
    \centering
    \includegraphics[width=0.9\linewidth]{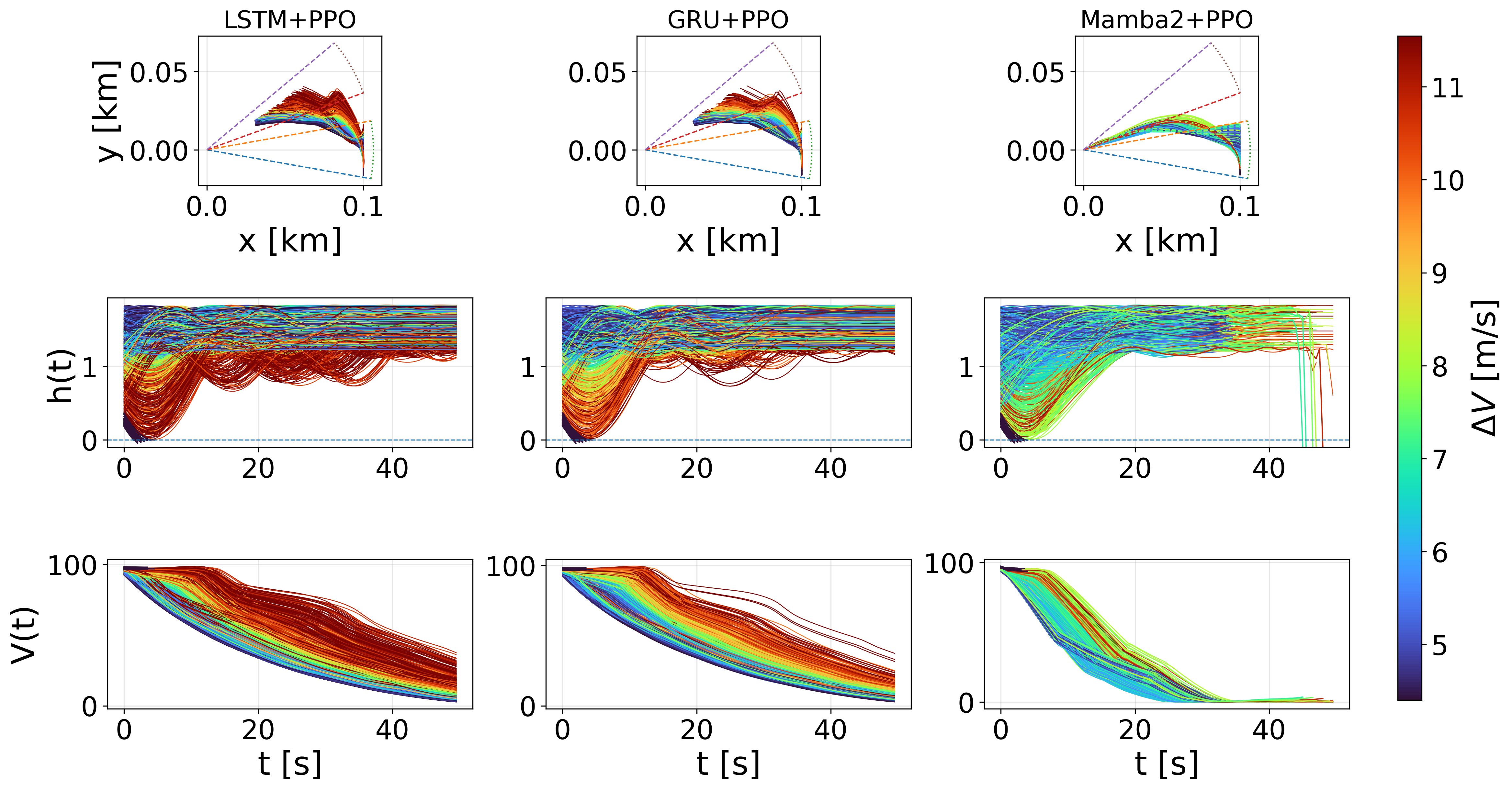}
    \caption{PPO docking trajectories}
    \label{fig:docking_ppo_traj}
\end{figure}

\begin{figure}[hbt]
    \centering
    \includegraphics[width=0.9\linewidth]{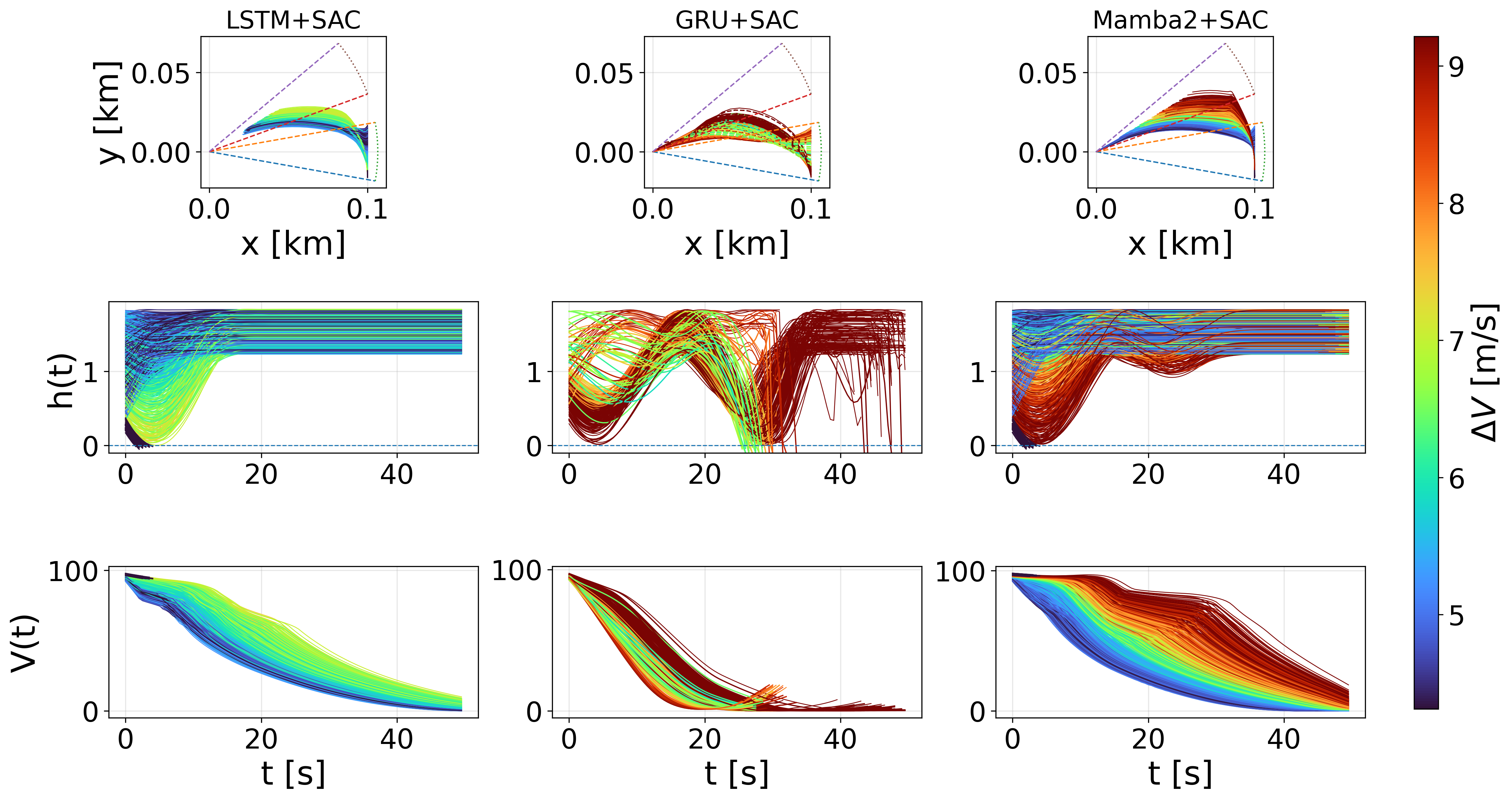}
    \caption{SAC docking trajectories}
    \label{fig:docking_sac_traj}
\end{figure}

As seen in Table \ref{tab:coop_uncoop_summary} LSTM and GRU models achieve the highest safety metrics but a $0.0\%$ dock rate, suggesting that the agents failed to learn how to reach the docking target within the given training time. Conversely, Mamba2+PPO achieves a $95.0\%$ successful dock rate while maintaining a $97.9\%$ safety margin. The lowest median fuel consumption is observed for LSTM+SAC, however at a docking success rate of 0\%. Instead, Mamba2 offers the second lowest median fuel consumption rate, with much higher rate of docking success.  Figure \ref{fig:docking_ppo_traj} evaluates the spacecraft docking scenario under nominal conditions for three PPO-trained architectures: LSTM+PPO, GRU+PPO, and Mamba2+PPO.  The top row shows planar $XY$ position trajectories of the chaser spacecraft, overlaid with a rotating docking-cone obstacle at its start and end orientations to illustrate the time-varying geometry of the safety constraint. The middle row plots the CBF value $h(t)$ over time, where $h < 0$ indicates a constraint violation, with the $h = 0$ boundary clearly marked. The bottom row shows the Lyapunov function $V(t)$. Analogous to the PPO docking group,  Figure \ref{fig:docking_sac_traj}  evaluates the standard spacecraft docking scenario for three SAC-trained architectures.


\subsection{Inspecting a Cooperative Target}
\begin{figure}[hbt]
    \centering
    \includegraphics[width=0.8\linewidth]{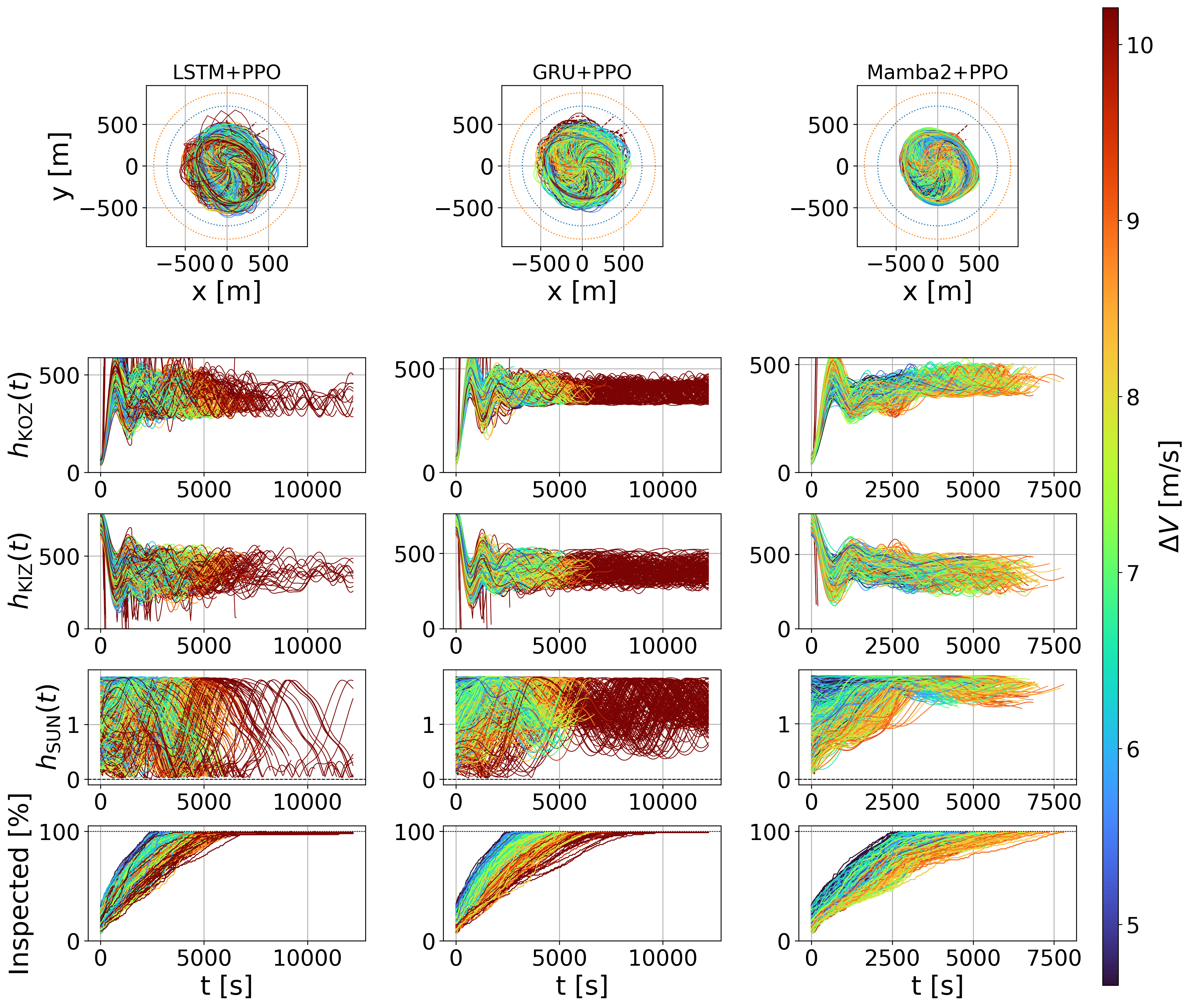}
    \caption{PPO inspection trajectories}
    \label{fig:inspection_ppo_traj}
\end{figure}

\begin{figure}[hbt]
    \centering
    \includegraphics[width=0.8\linewidth]{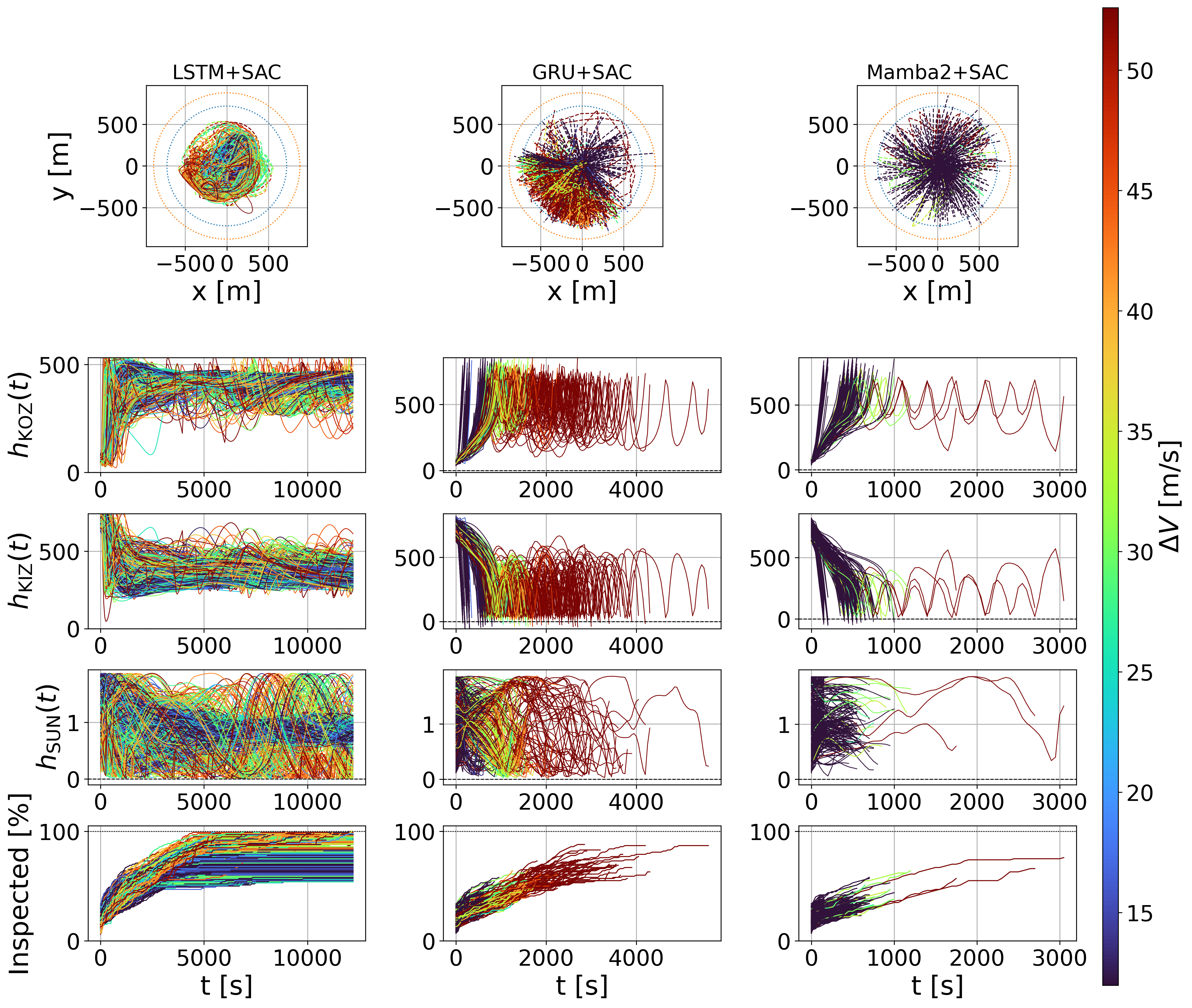}
    \caption{SAC inspection trajectories}
    \label{fig:inspection_sac_traj}
\end{figure}

The Inspection scenario pushes the agents further, requiring them to balance goal-oriented navigation against fuel efficiency and strict safety constraints. Figures~\ref{fig:inspection_ppo_traj} and \ref{fig:inspection_sac_traj} show the PPO- and SAC-trained network architectures, with one architecture per column and rows giving the $XY$ trajectories over the KIZ annulus, the KOZ barrier $h_\mathrm{KOZ}(t) = \|\mathbf{r}_b\| - (R_C + R_D)$, the KIZ barrier $h_\mathrm{KIZ}(t) = R_{\max} - \|\mathbf{r}_b\|$, the Sun-exclusion barrier $h_\mathrm{SUN}(t)$, and the cumulative inspection coverage. Table~\ref{tab:coop_uncoop_summary} reveals a safety collapse for SAC: both GRU+SAC and Mamba2+SAC record a $0.0\%$ safety rate. Among the PPO variants, GRU+PPO reaches high coverage ($98.11$ points) but inefficiently, consuming far more fuel (Thrust $\mu = 1042.96$) at only $67.8\%$ safety. Mamba2+PPO is again the best pairing, achieving near-complete coverage ($99.55\%$ median) and the highest safety ($99.4\%$) with a stable, efficient thrust profile ($\sigma = 115.44$).

 \subsection{Docking with an Adversarial Target.}

To assess robustness under unpredictable, actively evasive conditions, the models were evaluated on adversarial variants of the docking and inspection scenarios; given PPO's superior cooperative performance, only the PPO variants were carried forward. 

\begin{figure}[hbt]
    \centering
    \includegraphics[width=0.9\linewidth]{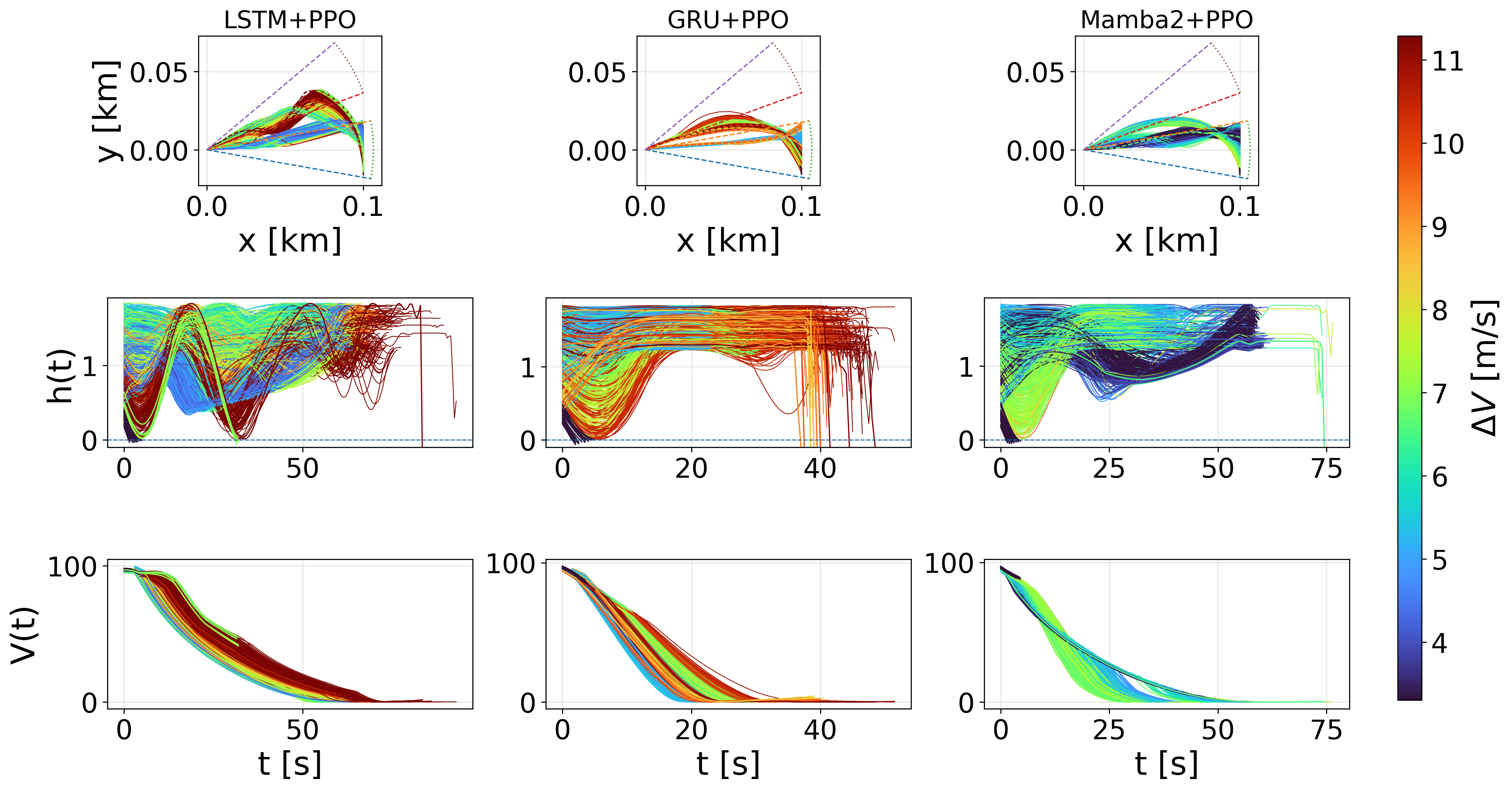}
    \caption{PPO adversarial docking trajectories}
    \label{fig:adversarial_docking_ppo_traj}
\end{figure}

For adversarial docking, the GRU backbone degrades sharply, falling to an $88.3\%$ safety rate and a $74.3\%$ docking-success rate. LSTM+PPO and Mamba2+PPO remain robust, with comparable docking success ($\sim 95\%$) and safety ($\sim 95.5\%$). Mamba2+PPO is the most fuel-efficient of the three, reaching the target with lower mean $\Delta v$ ($5.18$ vs.\ $6.77$ m/s) and lower variance ($1.87$ vs.\ $2.92$), indicating a more efficient and stable policy learned from fewer interactions. Figure~\ref{fig:adversarial_docking_ppo_traj} shows the adversarial docking trajectories for the three PPO network architectures, with the layout mirroring the cooperative docking figure.

\subsection{Inspecting an Adversarial Target. }
\begin{figure}[hbt]
    \centering
    \includegraphics[width=0.9\linewidth]{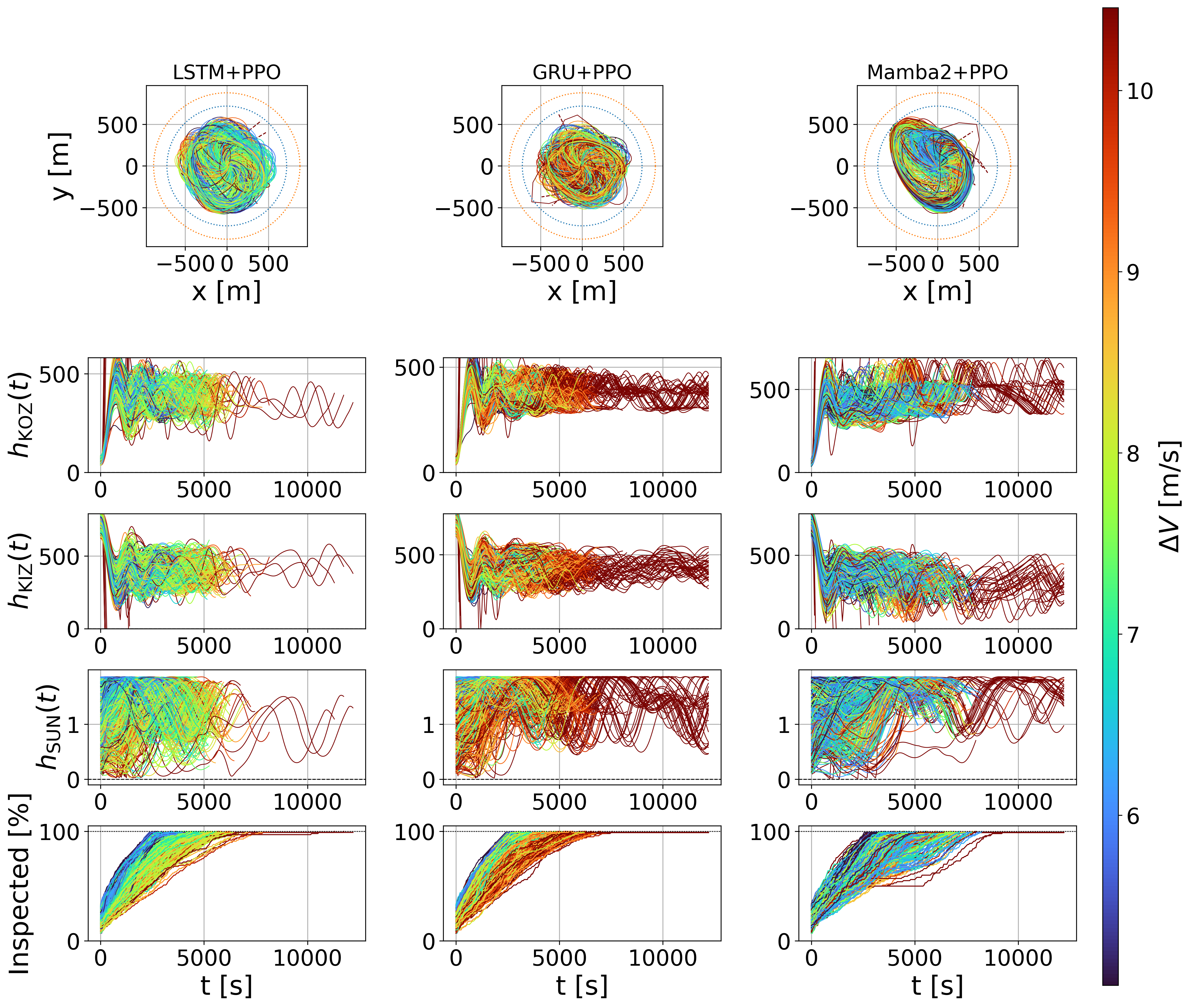}
    \caption{PPO adversarial inspection trajectories}
    \label{fig:adversarial_inspection_ppo_traj}
\end{figure}

Table \ref{tab:coop_uncoop_summary} demonstrates that all three models maintain excellent coverage against the adversarial target, inspecting roughly $98$--$99\%$ of the target points, but they rely on markedly different control strategies to
achieve this. The RNN-based models expend considerably more fuel countering the adversary's evasive maneuvers ($7.79$ and $9.69$~m/s mean $\Delta v$ for LSTM+PPO and GRU+PPO, respectively), and GRU+PPO does so while sacrificing safety
($67.8\%$ safe episodes). Mamba2+PPO, by contrast, dominates on every metric. It requires the least control effort ($5.29$~m/s) with the narrowest distribution ($\sigma = 1.37$~m/s), while simultaneously achieving the highest safety rate ($99.0\%$ vs.\ $95.6\%$ for LSTM+PPO) and the highest coverage ($99.4\%$). This across-the-board advantage highlights the superior
capacity of selective state space models for handling complex, multi-objective control problems, even in hostile environments.
Figure \ref{fig:adversarial_inspection_ppo_traj} presents results for the adversarial inspection scenario evaluated on the three PPO-trained architectures. The figure layout mirrors the standard inspection figure.

\section{Discussion}

Three observations emerge from these results. First, the performance gap between configurations scales with task difficulty. In Cruise Control all pairings are separated only by fuel, but in docking only Mamba2 learns to complete the task, in inspection the SAC variants collapse, and under adversarial behavior Mamba2+PPO consumes roughly $30$--$45\%$ less fuel than the RNN-based policies while matching or exceeding their safety and coverage. This illustrates that architecture and algorithm choices must be validated according to the difficulty of the problem.  Second, Mamba2's advantage is plausibly rooted in its linear, input-gated hidden-state recurrence, which, unlike the saturating nonlinear updates of the LSTM and GRU, mitigates the vanishing-gradient problem over long horizons and allows the policy to infer hidden task parameters and adversarial behavior from the full observation history. Third, the systematic underperformance of SAC reflects a mismatch between entropy-regularized, off-policy learning and the thin feasible sets of safety-critical control: the maximum-entropy objective injects stochasticity that repeatedly drives the agent across tight constraint boundaries, while replayed transitions generated under earlier class-$\mathcal{K}$ parameterizations leave the value estimates stale with respect to the current safety filter. On-policy methods such as PPO that learn only from data generated by the current filter can avoid this problem.

\section*{CONCLUSION}
 
This paper extended a previously developed meta-RL framework for tuning the class-$\mathcal{K}$ functions of ICCBFs by benchmarking three recurrent network architectures (LSTM, GRU, and Mamba2) and two training algorithms (PPO and SAC) across cruise control, docking, and inspection scenarios, including adversarial variants in which the target actively degrades the chaser's safety or sensor coverage. Monte Carlo evaluation under hidden-parameter, state, and thrust uncertainty showed that the choice of sequence model and training algorithm is fairly decisive rather than incidental, and that the performance gap between configurations widens with task complexity. On-policy PPO consistently outperformed off-policy SAC, whose entropy-driven exploration and stale replayed transitions led to safety collapse in the docking and inspection tasks. Among the network architectures, Mamba2 paired with PPO provided the best overall balance: it was the only configuration to achieve a high docking success rate ($95.0\%$) while preserving safety ($97.9\%$), attained near-complete inspection coverage ($99.6\%$) at the highest safety rate ($99.4\%$), and retained this advantage under adversarial behavior, where it achieved the highest safety and coverage at $30$--$45\%$ lower fuel consumption than the RNN-based alternatives. These results indicate that selective state-space models are well suited to learned safety filters for proximity operations. Future work will include adapting off-policy methods to the non-stationary safe set induced by the learned filter, and validating the framework on flight-representative hardware.
 

\section{Acknowledgment}
Research was sponsored by the Department of the Air Force Artificial Intelligence Accelerator and was accomplished under Cooperative Agreement Number FA8750-19-2-1000. The views and conclusions contained in this document are those of the authors and should not be interpreted as representing the official policies, either expressed or implied, of the Department of the Air Force or the U.S. Government. The U.S. Government is authorized to reproduce and distribute reprints for Government purposes notwithstanding any copyright notation herein.

\section*{Use of Artificial Intelligence}
AI tools were used in the preparation of this manuscript to improve the clarity of the technical writing. All scientific content, including the problem formulation, theoretical development, interpretation of results, and conclusions, is the sole intellectual contribution of the authors. The authors reviewed and verified all AI-generated text, and take full responsibility for the accuracy and integrity of the work.

\clearpage

\appendix
\section*{Appendix: A}
\label{app:A}

\begin{algorithm}
\caption{Recurrent Meta-RL via PPO (On-Policy)}
\label{alg:recurrent_ppo}
\small
\textbf{Require:} Task $\mathcal{E}$; recurrent actor $\pi_\theta$, critic $V_\phi$;
rollout length $T$, burn-in $K$, epochs $E$, clip $\epsilon$, GAE $(\gamma, \lambda)$.

\begin{algorithmic}[1]
\State Initialise $\theta,\, \phi$;\; context $\mathcal{C} \leftarrow \mathbf{0}^{K \times N}$
\For{each PPO update}
    \State \textbf{// --- Phase 1: Rollout ---}
    \State $\bm{h}^a,\, \bm{h}^c \leftarrow \mathbf{0}$
    \For{$t = 1$ \textbf{to} $T$}
        \State $a_t,\, \log\pi_t,\, \bm{h}^a \leftarrow \pi_\theta(\bm{s}_t,\, \bm{h}^a)$
        \State $v_t,\, \bm{h}^c \leftarrow V_\phi(\bm{s}_t,\, \bm{h}^c)$
        \State $\bm{s}_{t+1},\, R_t,\, d_t \leftarrow \mathcal{E}.\textsc{Step}\bigl(\textsc{ICCBF-QP}(a_t, \bm{s}_t)\bigr)$
        \If{$d_t^{(i)} = \textbf{true}$}\; $\bm{h}^a_{[:,i]},\, \bm{h}^c_{[:,i]} \leftarrow \mathbf{0}$
        \EndIf
        \State Store $(\bm{s}_t, a_t, R_t, d_t, v_t, \log\pi_t)$ in $\mathcal{B}$
    \EndFor
    \State \textbf{// --- Phase 2: GAE ---}
    \State $v_{T+1} \leftarrow V_\phi(\bm{s}_{T+1}, \bm{h}^c)$;\; $\hat{A}_{T+1} \leftarrow 0$
    \For{$t = T$ \textbf{downto} $1$}
        \State $\hat{A}_t \leftarrow (R_t + \gamma v_{t+1} - v_t) + \gamma\lambda\,\hat{A}_{t+1}$
    \EndFor
    \State \textbf{// --- Phase 3: Update with Burn-In ---}
    \State $\mathcal{C} \leftarrow \mathcal{B}.\textsc{obs}_{T-K:T}$
    \hfill$\triangleright$ save tail as burn-in context for next update
    \For{epoch $e = 1$ \textbf{to} $E$}
        \For{each minibatch $\mathcal{M} \sim \mathcal{B}$}
            \State Run $\pi_\theta, V_\phi$ over $[\mathcal{C},\, \mathcal{M}]$ from $\bm{h} = \mathbf{0}$;
                   discard first $K$ outputs
            \State $\hat{A} \leftarrow (\hat{A} - \mu_{\hat{A}}) / \sigma_{\hat{A}}$
            \State $r \leftarrow \exp(\log\pi'_t - \log\pi_t)$
            \State $\mathcal{L} \leftarrow -\mathbb{E}\bigl[\min(r\hat{A},\;
                   \operatorname{clip}(r, 1{\pm}\epsilon)\hat{A})\bigr]
                   + c_V\mathcal{L}_V - c_\mathcal{H}\mathcal{H}$
            \State Update $\theta, \phi$ via $\nabla\mathcal{L}$;\; clip $\|\nabla\|_2 \le \delta_g$
            \If{$\widehat{\mathrm{KL}} > \delta_{\mathrm{KL}}$}\; \textbf{break}
            \EndIf
        \EndFor
    \EndFor
\EndFor
\end{algorithmic}
\end{algorithm}

\clearpage

\begin{algorithm}
\caption{Recurrent Meta-RL via SAC (Off-Policy)}
\label{alg:recurrent_sac}
\small
\textbf{Require:} Task $\mathcal{E}$; recurrent actor $\pi_\theta$,
twin critics $Q_{\phi_1}, Q_{\phi_2}$, target critics $\bar{Q}_{\phi_1}, \bar{Q}_{\phi_2}$;
chunk length $L$, burn-in $K$, Polyak $\tau$, entropy coef $\alpha$.

\begin{algorithmic}[1]
\State Initialise $\theta,\, \phi_{1:2},\, \bar\phi_{1:2} \leftarrow \phi_{1:2}$,
       $\log\alpha$;\; replay buffer $\mathcal{D}$ of capacity $M$
\State $\bm{h} \leftarrow \mathbf{0}$;\; collect $N_{\mathrm{start}}$ random transitions into $\mathcal{D}$
\For{each environment step}
    \State \textbf{// --- Collection ---}
    \State $a_t,\, \bm{h} \leftarrow \pi_\theta(\bm{s}_t, \bm{h})$
    \State $\bm{s}_{t+1}, R_t, d_t \leftarrow \mathcal{E}.\textsc{Step}\bigl(\textsc{ICCBF-QP}(a_t, \bm{s}_t)\bigr)$
    \If{$d_t = \textbf{true}$}\; $\bm{h} \leftarrow \mathbf{0}$
    \EndIf
    \State Store $(\bm{s}_t, a_t, R_t, d_t)$ in chunk collector;\;
           flush complete chunks of length $K{+}L$ to $\mathcal{D}$
    \If{step $\bmod$ \texttt{train\_freq} $= 0$ \textbf{ and } $|\mathcal{D}| \ge$ \texttt{batch\_size}}
        \For{\texttt{gradient\_steps} iterations}
            \State \textbf{// --- Burn-In Split ---}
            \State Sample batch of sequences $(K{+}L)$ from $\mathcal{D}$
            \State $\mathcal{C} \leftarrow \text{seq}_{:K}$;\;
                   $(\bm{s}, a, R, \bm{s}', d) \leftarrow \text{seq}_{K:}$
            \hfill$\triangleright$ first $K$ steps warm up the RNN; loss on last $L$
            \State \textbf{// --- Critic Update ---}
            \State $a', \log\pi' \leftarrow \pi_\theta(\bm{s}', d,\; \mathcal{C}^{\,+1})$
            \hfill$\triangleright$ context shifted by 1 for next-obs
            \State $y \leftarrow R + \gamma(1-d)\bigl[\min_i \bar{Q}_{\bar\phi_i}(\bm{s}', a') - \alpha\log\pi'\bigr]$
            \State $\mathcal{L}_Q \leftarrow \sum_{i=1}^{2}\mathrm{MSE}\bigl(Q_{\phi_i}(\bm{s}, a,\; \mathcal{C}),\; y\bigr)$
            \State Update $\phi_{1:2}$ via $\nabla_\phi\mathcal{L}_Q$;\; clip $\|\nabla\|_2 \le \delta_g$
            \State \textbf{// --- Actor Update ---}
            \State $\tilde{a}, \log\pi \leftarrow \pi_\theta(\bm{s}, d,\; \mathcal{C})$
            \State $\mathcal{L}_\pi \leftarrow \mathbb{E}\bigl[\alpha\log\pi - \min_i Q_{\phi_i}(\bm{s}, \tilde{a},\; \mathcal{C})\bigr]$
            \State Update $\theta$ via $\nabla_\theta\mathcal{L}_\pi$;\; clip $\|\nabla\|_2 \le \delta_g$
            \State \textbf{// --- Entropy Coefficient Update ---}
            \State $\mathcal{L}_\alpha \leftarrow \mathbb{E}\bigl[-\alpha(\log\pi + \bar{\mathcal{H}})\bigr]$;\;
                   update $\log\alpha$ via $\nabla\mathcal{L}_\alpha$;\;
                   clamp $\alpha \ge \alpha_{\min}$
            \State \textbf{// --- Polyak Update ---}
            \State $\bar\phi_i \leftarrow \tau\phi_i + (1{-}\tau)\bar\phi_i$
            \hfill$\triangleright$ soft update target critics
        \EndFor
    \EndIf
\EndFor
\end{algorithmic}
\end{algorithm}





\clearpage

\bibliographystyle{AAS_publication}   
\bibliography{references}   

\end{document}